\PassOptionsToPackage{dvipsnames,table}{xcolor}
\documentclass{IOS-Book-Article}     

\usepackage[normalem]{ulem}
\usepackage[disable]{todonotes}
\usepackage[dvipsnames,table]{xcolor}
\usepackage{fvextra}
\usepackage{needspace}
\DefineVerbatimEnvironment{prompt}{Verbatim}{fontsize=\scriptsize, breaklines=true,
  frame=single, framerule=0.3pt, rulecolor=\color{gray!50}, numbers=left, numbersep=4pt,
  highlightcolor=gray!18}
\usepackage{orcidlink}
\usepackage{amsmath}
\usepackage{amssymb}
\usepackage{mathtools}
\usepackage{graphicx}
\usepackage{subcaption}
\usepackage{makecell}
\usepackage{booktabs}
\usepackage{url}
\usepackage{multirow}
\usepackage{cleveref}
\usepackage[T1]{fontenc}
\usepackage{hyperref}

\newcommand{\gpbam}{\textsc{GPBam}}
\newcommand{\norag}{\textsc{No-RAG}}
\newcommand{\lawrag}{\textsc{Law-RAG}}

\newcommand{\checkthis}[1][]{%
  \textcolor{red}{[check%
  \if\relax\detokenize{#1}\relax
  \else: #1%
  \fi]}%
}
\begin{document}
\begin{frontmatter}
\title{\gpbam{}: A Retrieval-Aware German Public Law Exam Benchmark\thanks{This paper is an extended version of a paper accepted at the 39th International Conference on Legal Knowledge and Information Systems (JURIX 2026).}}
\runningtitle{\gpbam{}}

\author[A\dagger]{%
\fnms{Abdullah Al} \snm{Zubaer}\orcidlink{0009-0001-0842-7434}%
\thanks{Corresponding Author,
abdullahal.zubaer@uni-passau.de. \(^{\dagger}\)~These authors contributed equally to this work.}%
},
\author[B\dagger]{%
\fnms{Lorenz} \snm{Wendlinger}\orcidlink{0000-0001-9459-6244}%
},
\author[A]{%
\fnms{Simon Alexander} \snm{Nonn}\orcidlink{0009-0007-7173-5302}%
},
\author[A]{%
\fnms{Michael} \snm{Granitzer}\orcidlink{0000-0003-3566-5507}%
}
and
\author[A]{%
\fnms{Jelena} \snm{Mitrovi\'c}\orcidlink{0000-0003-3220-8749}%
}

\runningauthor{A. A. Zubaer et al.}

\address[A]{%
Universität Passau, Passau, Germany
}
\address[B]{%
Technische Hochschule Deggendorf, Deggendorf, Germany
}

\begin{abstract}
Legal reasoning in civil-law systems is not rule-lookup: the German state
examination asks for a structured opinion (\emph{Gutachten}) that names the
governing norms, applies them to the facts, and argues to a conclusion. We
introduce GPBam (\textbf{G}erman \textbf{P}ublic Law \textbf{B}ar-style Ex\textbf{am}),
81 public-law examination cases with reference solutions, paired with a
recitation task over 200 statutory provisions. 
Drawn from state examination preparation, the cases are broad and rarely repeat, and are not publicly available - we find no evidence of exact-solution memorisation in the models we evaluate, and keeping them private limits contamination.
Because each reference solution's cited norms form a retrieval gold set over the very instances the essays are graded on, GPBam scores retrieval and long-form generation \emph{jointly}, which makes it a benchmark for legal retrieval as well as for legal writing and lets us ask how much retrieval quality a legal opinion actually needs.
Across 32 open-weight and proprietary models evaluated without retrieval, scores span 4.81--76.42 out of 100. 
Supplying the law helps modestly: across seven models and eight retrieval configurations, retrieval performance varies nearly threefold and buys at most $4.1$ essay points, and nothing significant for three of the seven, while an oracle with $91\%$ gold-norm recall improves scores by $2$--$6$ points.
Essay quality strongly tracks prior knowledge: verbatim statute recall, measured in a separate task, correlates at  $\rho{=}0.92$ over 32 models.
\end{abstract}


\end{frontmatter}

\section{Introduction}
\label{sec:introduction}
Many established legal benchmarks use classification, multiple-choice,
or short-answer formats \cite{chalkidis-etal-2022-lexglue,NEURIPS2023_89e44582,fei-etal-2024-lawbench}.
This permits standardized automatic scoring, but does not represent how case-based tasks in
German legal education and the First State Examination assess legal reasoning.
A typical case task requires a complete legal opinion (\emph{Gutachten}):
identify the issues and governing norms, state the requirements of each norm,
apply them to the facts, and reach a conclusion. The same structure recurs over
the issues and sub-issues of the case. A correct final conclusion can therefore
conceal missing or incorrect intermediate analysis, which label-based benchmarks
do not measure
directly~\cite{chalkidis-etal-2022-lexglue,NEURIPS2023_89e44582,
fei-etal-2024-lawbench}. Evaluating the full reasoning document is necessary for
this examination setting~\cite{nagl2026bengerbenchmarkingllmsystems}.

Several recent benchmarks include open-ended legal tasks.
GreekBarBench asks
questions from five areas of the Greek bar examination, supplies chapter-level
statutory context, and requires citations to both articles and case
facts~\cite{chlapanis-etal-2025-greekbarbench}. LEXam contains 2{,}841
open-ended questions from law examinations and accompanies them with guidance
on the expected reasoning approach~\cite{fan2026lexambenchmarkinglegalreasoning}.
For German law, BenGER evaluates complete exam-style case solutions alongside
short doctrinal tasks, but disables retrieval and tool-use endpoints for the
evaluated systems~\cite{nagl2026bengerbenchmarkingllmsystems}. Retrieval-focused
benchmarks address a different stage: LegalBench-RAG measures whether a system
locates the relevant spans, not whether those spans improve a complete legal
opinion~\cite{pipitone2024legalbenchragbenchmarkretrievalaugmentedgeneration}.
These benchmarks therefore emphasize different stages of legal problem solving.
Their complementary designs \textit{motivate} evaluating statutory recall, retrieval,
and downstream use within the same document-level task.

We introduce \gpbam{} (\textbf{G}erman \textbf{P}ublic Law \textbf{B}ar-style Ex\textbf{am}),
a benchmark that scores statute retrieval and long-form legal writing on the same 81 state-exam-style public-law cases, drawn from the University of
Passau's State Examination Preparation Course in Public Law. Each case pairs a fact pattern (\emph{Sachverhalt}) with an extensively reasoned reference solution. The task is
document-level: a model receives the case and writes a complete German
legal opinion (\emph{Gutachten}), including a table of contents and an assessment of every
question raised. The reference solutions average 7{,}080 words, compared with
924 for the case facts, and contain substantially more statutory references. This asymmetry is part of the task: the model must
determine much of the governing law rather than extract it from the question.
The cases are also broad and rarely repeat: the topics their authors label are mostly unique to a single case, and one case raises a median of eight of them across a median of twelve law books.
Keeping the cases and reference solutions non-public also reduces the risk that they enter model-training corpora and contaminate the evaluation~\cite{sainz-etal-2023-nlp}.
A direct text-overlap analysis finds no evidence of exact-solution memorisation of the reference solutions (\Cref{sec:contamination_analysis}).

GPBam evaluates statutory access in two complementary ways. First, the \textit{Essay Writing} task is run without retrieval and with statutory context retrieved from federal and Bavarian statutes.
We compare retrieval strategies and context budgets, and include an oracle condition that supplies the statutes cited by the reference solution, separating retrieval failure from failure to use relevant provisions.
Second, \textit{Article Recitation} names a provision and asks the
model to reproduce its text. The task covers 200 provisions: 100 frequently
cited by the GPBam reference solutions and 100 frequently cited by German
courts. It removes issue spotting and subsumption, providing a separate measure
of closed-book statutory-text recall. Essays are scored against the reference
solution by three open-weight judges; we take their per-case median and
separately measure the overlap of statutory citations.

Across 32 models evaluated without retrieval, essay scores span 4.81--76.42 points on the 0--100 scale. Calibration controls anchor the scale empirically: the reference solutions score 100, whereas wrong-case solutions and fact-pattern echoes score 0. Essay score correlates with Article Recitation on the GPBam provisions at $\rho{=}0.92$, while correlation with citation overlap, measured against the reference solution, is $\rho{=}0.94$. 
Supplying statutes via retrieval system has a limited effect in the present setup. 
Across the tested retrieval configurations, gold-norm recall rises nearly threefold, from 3.50\% to 9.61\% at $k{=}10$ and from 8.52\% to 23.92\% at $k{=}50$, while the largest non-oracle improvement over no retrieval is 4.1 essay score points, and for three out of seven generators no configuration improves significantly on no retrieval at all.
The oracle, which supplies the statutes the reference solution cites and reaches $91.31\%$ recall, raises essay scores by $2$--$6$ points and does so for every one of the seven evaluated models.
These results do not show that statutory knowledge causes higher essay scores. 
They \textit{do} show that supplying statutory text under the tested conditions does not close the observed performance gap, while closed-book statutory recall covaries strongly with essay quality. 
Our contributions are the following:

\begin{itemize}
  \item We introduce GPBam, 81 German state-exam-style public-law cases with
  reference solutions, formulated as complete legal-opinion writing rather than
  classification or short-answer prediction, and held out rather than published:
  we find no evidence of exact-solution memorisation of the reference solutions in
  any evaluated model, and keeping the material private limits contamination.
    \item Because each reference solution's cited norms form a retrieval gold set over the same 81 instances the essays are scored on, GPBam measures retrieval and long-form generation \textit{jointly}.
    This is what lets us ask how much retrieval quality a legal essay actually needs, and it makes the benchmark usable for legal retrieval independently of the essay task.
  \item We pair the essay task with Article Recitation over 200 provisions,
  making statutory recall observable separately from issue spotting,
  subsumption, and legal writing.
  \item We report a 32-model no-retrieval study using a three-judge ensemble, citation overlap, together with an analysis of inter-judge agreement.
  \item We evaluate statute retrieval for complete opinions across seven models,
  with controlled changes to retrieval method, corpus, and context budget, and
  an oracle upper bound using the reference solution's cited statutes.
    \item Findings show that supplying statutes helps modestly: a nearly threefold spread in gold-norm recall is worth at most $4.1$ points of essay score, with no significant gain for three of the seven generators, while an oracle at $91.31\%$ recall raises every generator by $2$--$6$ points.
\end{itemize}
\section{Related Work}

\subsection{From Legal Understanding to Open-Ended Examination Tasks}
\label{sec:rw_legal_benchmarks}

Many early legal NLP benchmarks framed the domain as classification, multiple-choice,
or other short/task-specific outputs.
LexGLUE groups document and clause classification
tasks~\cite{chalkidis-etal-2022-lexglue}, LegalBench assembles hand-built tasks
around common-law reasoning~\cite{NEURIPS2023_89e44582}, and LawBench organises
Chinese legal tasks into memorisation, understanding and
application~\cite{fei-etal-2024-lawbench}. Their evaluations are 
often outcome-focused, so a model
can be credited for a correct answer reached by faulty reasoning.
Prompting a general LLM on legal argument mining task did not by itself
dissolve these tasks~\cite{10.3389/frai.2023.1278796}.
Nearer to the present setting, open-weight LLMs have been put to a range of
German legal-education tasks, where capability falls away as the material turns
doctrinal: labelling the components of
\emph{Gutachtenstil}, and working over whole legal opinions rather than short
spans. On those two, neither a longer context window nor prompt engineering
carried the models past a bag-of-words
baseline~\cite{10.1007/978-3-032-21477-5_11}.

Recent benchmarks increasingly include open-ended examination tasks. Outside law,
SciEx grades free-text exam answers with both human experts and automatic
graders~\cite{dinh-etal-2024-sciex}. Within law, \textsc{oab-bench} draws 105
questions from the Brazilian bar examination and applies the official grading
guidelines through an LLM judge~\cite{10.1145/3769126.3769227}. GreekBarBench
covers five areas of the Greek bar and requires citations to statutory articles
and to case facts, scored on three
dimensions~\cite{chlapanis-etal-2025-greekbarbench}.
LEXam derives 7{,}537 questions from 340 Swiss law exams, 2{,}841 of them open-ended. Each
open question is paired with a reference answer and reasoning approach expected~\cite{fan2026lexambenchmarkinglegalreasoning}.   
PLawBench and JurisBench extend the setting to professional practice in Chinese
law, the former with expert-written rubric items across 13
scenarios~\cite{shi-etal-2026-plawbench}, the latter through a multi-stage
litigation workflow in which errors
cascade~\cite{chen-etal-2026-jurisbench}. Surveys of the area report the same
trajectory~\cite{HE2026100906}.

Two properties separate this line of work from GPBam. The first is the unit of
evaluation. LEXam, PLawBench and JurisBench score answers to individual
questions or workflow stages, and
GreekBarBench scores one answer to a case-specific question against
span-based rubrics. GPBam scores one complete \emph{Gutachten} per case: a table
of contents followed by a full assessment that answers every question raised in
the \emph{Sachverhalt}, with no study-imposed output-token cap. 
The reference solutions are correspondingly long, at $7{,}080$ words on average against $924$ for the facts - roughly twenty-nine times the $247$-word mean reference answer of LEXam~\cite{fan2026lexambenchmarkinglegalreasoning}, whose open questions are individually scorable sub-problems rather than whole opinions. 
GreekBarBench scores one answer per question against expert-annotated spans for facts, cited articles and analysis~\cite{chlapanis-etal-2025-greekbarbench}, so its unit is a passage within an answer. 
The biggest difference is jurisdiction. Common-law and Chinese material dominates, while civil-law systems, in which the analysis is anchored in codified statutes rather than precedent, are less well covered~\cite{fan2026lexambenchmarkinglegalreasoning}. 
For German law, employment-contract clauses have been classified as valid, unfair or void~\cite{10.1145/3769126.3769209}, which isolates subsumption at clause level. 

Among the benchmarks reviewed, BenGER is the closest work to ours.
It evaluates 596 exam-style cases spanning civil, criminal, and public law and multiple stages of legal education, together with 531 shorter doctrinal items.
Its judge scores ten weighted dimensions and is checked against three independent blind human gradings on a controlled subset~\cite{nagl2026bengerbenchmarkingllmsystems}. Retrieval and tool-use endpoints are switched off throughout, so its systems answer from parametric knowledge alone.
GPBam is narrower - 81 state-exam-style public-law cases from a single state-examination preparation course - and uses that depth for two things BenGER does not measure: what happens when the statutes are supplied via retrieval, and how
much of the statutory text models retain without external knowledge analogues to closed-book performance.

\subsection{Legal Context and LLM-Based Evaluation}
\label{sec:rw_benchmark_evaluation}

\paragraph{Supplying legal context.} 
Retrieval augmentation conditions  generation on passages fetched at inference time~\cite{lewis2020rag}, and the  pipelines now in use extend that basic loop on both sides of the retriever -  rewriting or diversifying the query before search~\cite{ma-etal-2023-query,  gao-etal-2023-precise}, fusing a dense and a lexical channel~\cite{cormack2009rrf,  robertson2009bm25}, and re-ranking the pool before it reaches the generator -  a family the surveys group as \emph{advanced} and \emph{modular}  RAG~\cite{gao2024ragsurvey}. Our retrieval ladder walks exactly these components over statutory law.
The central question, however, is not only whether these techniques retrieve more relevant legal material, but whether those retrieval gains improve the model's final answers.
A relationship that legal RAG evaluations have only recently begun to examine directly.

Much of the existing legal literature evaluates retrieval separately from generation.
LegalBench-RAG, for example, scores the retrieval step alone~\cite{pipitone2024legalbenchragbenchmarkretrievalaugmentedgeneration}. 
Pairing a statutory corpus with a labour-law benchmark puts retrieval-augmented accuracy well below the claims made for it~\cite{10.1145/3769126.3769256}. 
An evaluation of three commercial legal research tools built on retrieval measures hallucination rates between 17\% and 33\%~\cite{https://doi.org/10.1111/jels.12413}.
Varying the German legal context supplied to the model shows that full-text
sources help only moderately, while distilled examination guidelines help
more~\cite{10.1145/3769126.3769209}. Context length can reduce performance even under perfect retrieval and when
irrelevant tokens are masked~\cite{du-etal-2025-context}. Where retrieval and downstream answering are evaluated jointly, better access
to relevant legal context does not necessarily yield better answers. In Bar
Exam QA, which uses short, stylized fact patterns with four answer choices, and
Housing Statute QA, which uses binary yes/no questions, structured
legal-reasoning query expansion improves recall for several retrievers, but
these gains do not consistently translate into higher answer
accuracy~\cite{10.1145/3709025.3712219}.

These studies therefore either evaluate retrieval in isolation or measure its
downstream effect on question-answering or classification tasks. None of the
reviewed studies examines whether retrieved statutory text improves a complete
examination essay in a civil-law system. GPBam runs the same 81 cases with and
without retrieval and adds an oracle condition that supplies the statutes cited
by the reference solution, which separates ``retrieval cannot find the norms''
from ``the norms do not help''.

\paragraph{Probing statutory knowledge directly.} A parallel line measures
whether models hold statutory text at all. LawBench includes a memorisation
level~\cite{fei-etal-2024-lawbench}; CLaw segments 306 Chinese national statutes
to subparagraph level and reports that models struggle to reproduce provisions
faithfully~\cite{xu2025clawbenchmarkingchineselegal}; LeKUBE evaluates statute recitation and recall
under knowledge updates~\cite{10.1145/3673791.3698407}. GPBam adopts the same idea as a
separate closed-book task over 200 provisions, and then correlates it with essay
quality across models. This is what lets us ask whether the bottleneck on a
\emph{Gutachten} is knowing the law or applying it.

\paragraph{LLM judges.} Judging long-form legal answers at benchmark scale is
constrained by the cost of expert annotation: it is slow, requires specialised
lawyers, and does not scale to the number of models and conditions a benchmark
must cover~\cite{chlapanis-etal-2025-greekbarbench,enguehard-etal-2025-lemaj}.
The expert grades themselves are also noisy. Where the same law-exam answers
have been regraded by many markers, the harshest and the most generous mark on
one answer differ by $6.47$ points on average, on a scale that runs from $0$ to
$18$~\cite{2024_hufeld_jede_korre,nagl2026bengerbenchmarkingllmsystems}.
This variability cautions against treating one marker's grade as definitive ground truth.
Lexical overlap with a reference solution is no substitute~\cite{fan2026lexambenchmarkinglegalreasoning}, because legally equivalent
analyses differ in wording. Recent benchmarks therefore grade with LLM
judges~\cite{li-etal-2025-generation,chlapanis-etal-2025-greekbarbench,fan2026lexambenchmarkinglegalreasoning,nagl2026bengerbenchmarkingllmsystems}.
Reported failure modes include self-preference, where an evaluator favours its
own generations~\cite{panickssery2024llmevaluatorsrecognizefavor}, and
sensitivity to prompt~\cite{GU2026101253}. Benchmarks respond by
validating the judge: GreekBarBench builds a meta-evaluation set and finds that
span-based rubrics improve
alignment~\cite{chlapanis-etal-2025-greekbarbench}; LEXam uses a judge ensemble
with expert validation~\cite{fan2026lexambenchmarkinglegalreasoning}; BenGER benchmarks its
judge against a pool of human markers, obtaining $r{=}0.76$ under Pearson and
$\kappa{=}0.60$~\cite{nagl2026bengerbenchmarkingllmsystems}; \textsc{oab-bench} reports
strong correlation between a frontier judge and human
examiners~\cite{10.1145/3769126.3769227}. A parallel line changes the unit of
assessment instead of validating a holistic one: LeMAJ
splits a legal answer into small,
individually checkable units it calls \emph{legal data points}, and scores those
without consulting a reference answer~\cite{enguehard-etal-2025-lemaj}, and DeCE separates precision from
recall against criteria extracted from the gold answer, reporting a correlation
with expert judgement of $r{=}0.78$ on a legal question-answering task with
citation grounding, against $r{=}0.35$ for pointwise
scoring~\cite{yu-etal-2025-beyond-pointwise}. For German exam scripts in
particular, a comparison of 27 models against examiner marks under quadratic
weighted kappa finds that pooling several models tracks the examiners more
closely than any single model does, by as much as $0.15$
kappa~\cite{zubaer2026gradelegalautomatedgradinggerman}.
Within GradeLegal's two datasets, agreement  was far
higher on public-law scripts than on criminal-law ones, so the domain
GPBam covers is the one where automatic marking has so far proved most
tractable. GPBam scores each essay against the reference solution with three open-weight
judges and takes the per-case median. We report inter-judge agreement at essay and model
level and report a fitted adjustment for judge-generator self-preference.
\section{The GPBam Benchmark}

\subsection{Examination Setting}
\label{sec:german_exam}
German public-law examination cases present a fact pattern and one or more questions whose solution often requires procedural and substantive legal analysis.
The expected solution is not a list of rules but a structured legal opinion (\textit{Gutachten}).
Following the legal opinion style (\textit{Gutachtenstil}), the reasoning typically proceeds through four components: an \textit{Obersatz} identifying the legal issue to be examined, a \textit{Definition} specifying the relevant legal requirements, the \textit{Subsumtion} of the facts under those requirements, and an \textit{Ergebnis} stating the resulting conclusion.
This structure recurs across relevant issues and sub-issues in the solution, making the appropriate unit of evaluation an entire reasoning document rather than an isolated answer span.

The source documents of GPBam belong to the \emph{Examenskurs im \"{O}ffentlichen Recht} (State Examination Preparation Course in Public Law) of the University of Passau's Institute for Legal Didactics. They are practice examination written at the level and in the format of the German state examination.
We therefore use ``state-exam style'' rather than treating the documents as official state-examination papers. Each case fact pattern is accompanied by an extensively reasoned model solution that serves as the reference solution for the respective case.

\subsection{Dataset Construction and Characteristics}
\label{sec:dataset_construction}

\textsc{GPBam} comprises 81 German public-law examination cases sourced from \emph{Examenskurs im \"{O}ffentlichen Recht} (State Examination Preparation Course in Public Law) of the University of Passau's Institute for Legal Didactics. Each case pairs a \emph{Sachverhalt}, a factual scenario followed by the questions the examinee must answer, with a \emph{L\"{o}sung}, a reference solution
that develops the expected legal analysis in full.

\paragraph{Length.}

\Cref{tab:dataset-stats} summarizes the corpus. The two components are strongly asymmetric: reference solutions are around $8\times$ longer than case facts in words ($7080$ vs. $924$) and in tokens ($13855$ vs. $1657$). The task is therefore long-form generation, rather than the short-answer and multiple-choice formats of many widely used legal benchmarks~\cite{NEURIPS2023_89e44582,chalkidis-etal-2022-lexglue}, placing it alongside with recent
free-text legal benchmarks~\cite{fan2026lexambenchmarkinglegalreasoning, chlapanis-etal-2025-greekbarbench}.

\begin{table}[htbp]
\centering
\footnotesize
\setlength{\tabcolsep}{1.5pt}
\caption{
Statistics of the \textbf{GPBam} corpus. 81 case--solution pairs, 1.3M tokens.
Values are the mean over instances with the sample standard deviation as a subscript, rounded to integers.
\S-refs is the number of \emph{unique} (law book, \S{} or Art.) statutory references per document, counting both \S- and Art.-form citations,
\textbf{Books} the number of distinct law books those references fall in, and \textbf{Topics}
the number of entries in the \emph{Themen des Falles} list the course authors append to each
solution (present in 80 of the 81; case facts carry none).
}
\begin{tabular}{@{}lrrrrrr@{}}
\toprule
\textbf{Component} & \textbf{\#Docs} & \textbf{Words} & \textbf{Tokens} & \textbf{\S-refs} & \textbf{Books} & \textbf{Topics} \\
\midrule
Case facts (\emph{Sachverhalt}) & 81 & $924_{\pm 293}$ & $1657_{\pm 508}$ & $4_{\pm 4}$ & $3_{\pm 2}$ & -- \\
Reference solution (\emph{Lösung}) & 81 & $7080_{\pm 1789}$ & $13855_{\pm 3610}$ & $52_{\pm 16}$ & $12_{\pm 3}$ & $9_{\pm 2}$ \\
\addlinespace[2pt]
\multicolumn{7}{@{}l@{}}{\textit{Range (min--max)}} \\
\quad Case facts & & 272--1904 & 469--3103 & 0--13 & 0--10 & -- \\
\quad Reference solution & & 2989--10917 & 5788--22721 & 20--97 & 5--20 & 3--15 \\
\midrule
\multicolumn{7}{@{}l@{}}{Language: German\quad$\cdot$\quad Domain: Public law (\emph{Öffentliches Recht})} \\
\multicolumn{7}{@{}l@{}}{Source: State Examination Preparation Course in Public Law, University of Passau} \\
\multicolumn{7}{@{}l@{}}{Aligned case--solution pairs: 81\quad$\cdot$\quad Corpus size: 1.3M tokens} \\
\multicolumn{7}{@{}l@{}}{Topics: 520 distinct; $83\%$ occur in one case, none in more than 13} \\
\bottomrule
\end{tabular}

\label{tab:dataset-stats}
\end{table}

{\color{black}
\paragraph{Legal density.} Length alone understates the difficulty. Case facts
cite almost no law ($4_{\pm 4}$ unique statutory references), whereas the
corresponding reference solutions cite $52_{\pm 16}$. This gap characterises the
core challenge: models must identify the governing law themselves rather than
apply law supplied to them. It motivates both experimental conditions: supplying
the relevant statutes through retrieval
(\Cref{sec:generation_conditions}), and probing statutory knowledge directly
through article recitation (\Cref{sec:article_recitation_task}).
}

\paragraph{Topical coverage.} The cases are written as preparation for the
state examination, therefore they are set to span the examinable material rather than
to focus on one doctrine. Each reference solution closes with a list of the topics
its case examines, compiled by the course authors; 80 of the 81 solutions carry
one. These lists name 520 distinct topics, of which $83\%$ appear in exactly one
case, and the most frequent of them, the \emph{Anfechtungsklage} (action for annulment), in 13.
A single case raises a median of eight topics and cites a median of twelve
distinct law books, so breadth is a property of the individual case and not only
of the collection.

\paragraph{Contamination analysis.}
\label{sec:contamination_analysis}
A case whose solution a model had already read would be answered from memory rather than from reasoning.
We probe this on the essays directly: for each, we measure the longest exact word-span it shares with its own reference solution, and the same quantity against four other control cases' solutions, which share the genre, the \emph{Gutachtenstil} conventions and much statutory wording but nothing case-specific.
Memorisation of a \emph{L\"{o}sung} would show as spans of dozens to hundreds of words above that control; it does not appear.
The median longest match is 7 words against a case's own solution and 5 against the controls, an excess of two; the longest raw match is 38 words and the largest excess over control 34, with only five essays at twenty or above.
Per-model median excess ranges from one to three words across all 32 generators, including the two highest-scoring systems, whose longer raw matches are offset by equally longer control matches.
All five are quoted source material rather than recalled analysis - the question posed in the \emph{Sachverhalt}, passages of statutes, and a court passage that essay and solution both cite.
The test only bounds the strongest exposure, to the exact reference solutions, but does not test for paraphrases or the \emph{Sachverhalt} alone.

\subsection{Benchmark Tasks}
\label{sec:benchmark_tasks}

\begin{figure}[htbp]
    \centering
    \includegraphics[width=\linewidth]{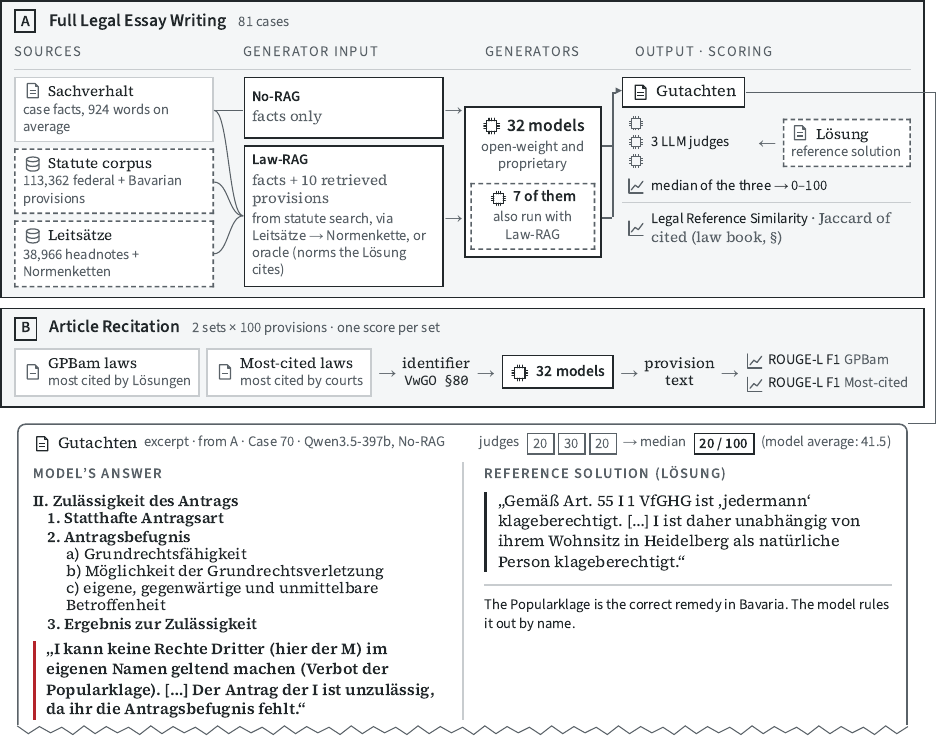}
    \caption{\textbf{The two \gpbam{} tasks and their data.} Essay writing gives a model one \emph{Sachverhalt} and asks for a complete \emph{Gutachten}, with and without retrieved context; Article Recitation names a provision and asks for its text, with no case and no retrieval. The reference solution is only used as the judges' target, and the norms it cites are both the retrieval gold set and the \emph{\gpbam{} laws} provision set. The excerpt illustrates one case.}
    \label{fig:gpbam-tasks}
\end{figure}

\Cref{fig:gpbam-tasks} summarizes the GPBam benchmark tasks, which separates integrated legal writing from statutory recall and consists of two distinct but related tasks: (1) full legal essay writing and (2) article recitation.
Essay writing uses 81 case/solution pairs of~\Cref{sec:dataset_construction} and is run in two context conditions, without and with retrieved context; besides that context, the two conditions also differ in the framing of the instruction (\Cref{sec:essay_task}).
Article recitation uses two fixed sets of 100 provisions, involves no fact pattern, and is scored against official statutory text.
The first task measures what a model produces when the applicable law has to be identified and applied.
The second measures how accurately it reproduces statutory wording when the provision is named for it.~\Cref{tab:task_summary} provides further details.

\subsubsection{Full Legal Essay Writing}
\label{sec:essay_task}

\paragraph{Task.} The input is the case facts to the model and the expected output is the complete legal opinion in German that answers every question raised in the case facts.
The reference solution is withheld during generation and is never part of the models' input, it serves as the evaluation target used by the judge ensemble (\Cref{sec:judges},~\Cref{sec:metrics}).

\paragraph{Instruction.} All 81 cases are inserted into a German instruction
template with two slots, ``context'' and ``case text''. The template describes
the input as a state-examination-level case to be answered within four hours and
asks for a table of contents followed by a complete legal assessment. It
requires citations to identify the paragraph (Abs.), sentence (S.), number
(Nr.), or letter (Buchst.) of a provision where applicable, and instructs the
model to avoid unspecific references. The prompt specifies no output-length
limit. \norag{} and \lawrag{} use variants of the same template, with the
differences described under \emph{Conditions} below. See the complete prompts
in Appendix~\ref{appendix:qa_prompt} and~\ref{appendix:qa_rag_prompt}.

\paragraph{Conditions.} In \norag{} the context slot is empty, so the case
facts are the only case-specific input. In \lawrag{} the slot holds a
\texttt{<context>} block containing the retrieved provisions, each wrapped in a
\texttt{<norm>} element that carries its citation, its jurisdiction and its
heading as attributes, so that the model can cite the provision rather than its
position in the list.

The case text is identical across the two conditions, but the instruction is
not. The \lawrag{} template adds guidance on how to read the retrieved
material: that the selection is incomplete and partly irrelevant, that state law
may be missing from it even where it governs, that a provision's absence from
the block is not evidence that it does not apply, and that the essay should cite
the applicable norms whether or not their text is printed. It also drops three \norag{}-only sentences, including an instruction to cite
exclusively from the considerations the model has identified. The two conditions therefore differ in
access to statutory text \emph{and} in the instruction that frames it, which is
a confound to keep in view when reading any \norag{}--\lawrag{} difference.

\begin{table}[htbp]
\centering
\caption{\textbf{GPBam tasks and data.} Essay writing measures integrated legal
analysis under two context conditions; Article Recitation isolates verbatim
statutory knowledge in a single closed-book condition. Word counts are means
with sample standard deviation (\Cref{tab:dataset-stats}); the reference
solution is never part of the model input.}
\label{tab:task_summary}
\small
\resizebox{0.9\linewidth}{!}{%
\begin{tabular}{@{}
  >{\raggedright\arraybackslash}p{0.165\textwidth}
  >{\raggedright\arraybackslash}p{0.405\textwidth}
  >{\raggedright\arraybackslash}p{0.375\textwidth}@{}}
\toprule
 & \textbf{Full Legal Essay Writing}
 & \textbf{Article Recitation} \\
\midrule
Ability probed
 & Issue spotting, rule selection, subsumption, structured argument
 & Recall of statutory wording \\
Input
 & Case facts (\emph{Sachverhalt}) and the questions posed
 & Provision identifier only (e.g.\ \mbox{\texttt{VwGO \S~154}}) \\
Context conditions
 & \norag{} (facts only) and \lawrag{} (facts $+$ retrieved context: ten
   passages by default, fifty in the $k{=}50$ arms, headnote text in the LS-text
   arm, and the full text of all provisions cited by the solution in the oracle;
   \Cref{tab:ladder})
 & One condition; no retrieval, no case facts \\
Output
 & Complete German \emph{Gutachten}: table of contents and full legal
   assessment; no length limit
 & Text of the named provision \\
\addlinespace[2pt]
Sets $\times$ items
 & $1 \times 81$ aligned case-solution pairs
 & $2 \times 100$ provisions\\
Item selection
 & \emph{Examenskurs im \"{O}ffentlichen Recht}, University of Passau
  & \emph{GPBam laws}: 100 most-cited provisions (\S\ and Art.) of the 81 solutions;
   \emph{Most-cited laws}:  100 most cited per the Open Legal Data citation network from\cite{milz2021analysis} \\
Evaluation target
 & Reference solution (\emph{L\"{o}sung})
 & Official statutory text (federal-law snapshot \texttt{67e0567}; graph
   records) \\
\addlinespace[2pt]
Primary metric
 & Median of a three-judge LLM ensemble, rescaled to 0--100
 & ROUGE-L F1 against the statutory text, without stemming \\
Secondary signal
 & Legal Reference Similarity: Jaccard overlap of unique (law book, \S/Art.) pairs
 & - \\
\addlinespace[2pt]
Prompts per model
 & 81 ($81$ cases, \norag{}); 810 for the seven models that also carry the
   retrieval ladder ($81$ cases $\times$ 10 settings: \norag{}, eight retrieval arms and the oracle;
   \Cref{tab:ladder})
 & 200 ($2$ sets $\times$ 100 provisions) \\
\midrule
\multicolumn{3}{@{}l@{}}{Language: German\quad$\cdot$\quad Domain: Public law
(\emph{\"{O}ffentliches Recht})} \\
\multicolumn{3}{@{}p{0.945\textwidth}@{}}{Population: the same 32 models receive
every \norag{} essay prompt and every Article Recitation prompt (\Cref{sec:models}; recitation exceptions in~\Cref{gpbam:para:scope});
\lawrag{} covers seven models (\Cref{tab:ladder})} \\
\bottomrule
\end{tabular}
}
\end{table}

\subsubsection{Article Recitation}
\label{sec:article_recitation_task}

\paragraph{Task.} Article Recitation asks a model to reproduce a single statutory provision from its identifier alone. The input is a law book and a number, as stored in the respective set. The expected output is the text of that provision.
The German prompt requests the exact wording of the current version, without
warnings, an introduction, or follow-up questions, and the most detailed
available summary if exact reproduction is not possible.
The generated text is compared with the stored official text of the provision using ROUGE-L F1 (\Cref{sec:metrics}). No case facts, retrieval, or reference solution are involved in this task. See the complete prompt in~\Cref{appendix:article_recitation_prompt}.

\paragraph{Provision sets.}

A provision is named the way a German lawyer cites one: by the \emph{law book} it
sits in - the statute, such as the Verwaltungsgerichtsordnung (VwGO) - and its number
within that statute. Each of the two sets holds 100 such provisions.

The \textit{GPBam laws} set comes out of the benchmark itself. We extract the
statutory citations of the 81 reference solutions in both their \S- and Art.-form
and order the provisions they cite
by how often they are cited, which brings the material the examiners lean on to
the top. Scoring an answer requires the provision's official text, so we look each
candidate up in the pinned \texttt{gesetze-im-internet} snapshot
(\Cref{sec:release}) and keep the 100 most-cited for which that lookup returns a
text. They come from 19 law books, led by the Verwaltungsgerichtsordnung (VwGO),
the Grundgesetz (GG) and the B\"urgerliches Gesetzbuch (BGB).

The \textit{Most-cited laws} set comes from outside the benchmark, and reflects
what German courts cite rather than what these examinations ask: the 100
provisions appearing in the largest number of decisions in the Open Legal
Data\footnote{\url{https://openlegaldata.io}} citation
network~\cite{milz2021analysis}. These come from nine law books, most of them the
Zivilprozessordnung (ZPO), the Verwaltungsgerichtsordnung (VwGO) and the
B\"urgerliches Gesetzbuch (BGB).

\paragraph{Scope.}
\label{gpbam:para:scope}
Article recitation removes issue spotting and subsumption by construction. It probes access to statutory wording; not legal analysis, and does not substitute for essay evaluation. ROUGE-L credits reproduction of the statutory text, therefore paraphrases and correct but abridged text receive partial credit at best. Every model answers the same 200 prompts\footnote{With two exceptions. Three of Gemini-3.7-flash's prompts returned an upstream serving error on every attempt, so its two scores are means over the remaining 99 and 98 provisions. Open\_steuerllm, EuroLLM-22B-I and Ministral-3-14B-R were not run on the 27 provisions that entered the \emph{\gpbam{} laws} set when it was re-derived with the \S/Art.\ extractor; their scores on that set are means over the other 73.}, which puts the two sets on one scale
(\Cref{tab:main_results}) and allows either to be related to essay quality at
model level (\Cref{sec:predictors}).

\subsubsection{Benchmark Records and Access}
\label{sec:access}
\label{sec:release}

Both halves of the law corpus are pinned: federal law to \texttt{gesetze-im-internet} revision \texttt{67e0567}\footnote{\url{https://github.com/QuantLaw/gesetze-im-internet/commit/67e0567}}, a scrape dated 2026-02-09 that parses to 106{,}354 provisions across 5{,}936 law books, and Bavarian state law to a crawl of BAYERN.RECHT's statute index taken on 2026-08-31, giving 7{,}008 articles from 720 statutes.

The cases and reference solutions are teaching material of the Institute for Legal Didactics.
Most are based on original state-examination papers, which places them under copyright, and they remain in active use in the examination course they were written for.
They are therefore not redistributed.
\gpbam{} is a held-out benchmark: we evaluate on it and release raw scores, but not the underlying documents or generations.
Scores, code and the retrieval task (query rewrites, statute corpus) are released at \url{https://github.com/wendli01/gpbam}

\label{sec:experimental_protocol}

\subsection{Model Selection}
\label{sec:models}

\Cref{tab:models} shows the models used in our experiments. We selected a heterogeneous candidate pool to cover dimensions plausibly related to long-form legal performance: capability, parameter scale, architecture, training objective, developer family, and deployment regime.
This breadth reduces the risk that the benchmark findings reflects one model family or configuration.
Because a complete legal opinion requires sustained, multi-step analysis, we include both standard instruction-tuned and explicitly reasoning-oriented models rather than restricting the evaluation to one inference paradigm.
We likewise include proprietary hosted systems and open-weight models.
The later permit on-premise deployment and processing, thus greater control over sensitive legal texts in contrast to proprietary systems~\cite{10.1145/3708821.3733888, huang-etal-2025-middle}.
Across these deployment settings, the model coverage spans dense and mixture-of-experts architectures and a wide range of parameter scales.
Beyond the general-purpose families, we include SteuerLLM~\cite{Wind_2026}, a German tax-law adapted model, to examine whether legal-domain adaptation transfers to public-law examination tasks, and two European models trained outside the major vendor families: the multilingual EuroLLM~\cite{ramos2026eurollm22btechnicalreport} and Soofi-S-Isar~\cite{soofiteam2026sovereignopensourcefoundationmodel}, a German sovereign-AI consortium release.

The main evaluation set contains 32 models: 29 from nine general-purpose
developer families, together with SteuerLLM, EuroLLM and Soofi-S-Isar.
Twenty-five are open-weight releases, all with published parameter counts
ranging from 3B to 685B; the remaining seven are proprietary hosted systems
whose providers disclose none. Among the 25 open-weight models, twelve use
dense architectures and thirteen use mixture-of-experts architectures with
3B--41B active parameters. The retrieval ladder covers seven generators.

\begin{table}[ht]
\centering
\caption{\textbf{Models used in the experiments.} Identifier is the string used to query
the serving endpoint; short name is the label used in the results tables. Size
is the parameter count published by the model provider, given as
total/active for mixture-of-experts models and as a single figure for dense
ones.
\emph{n/d} marks systems whose provider does not disclose a count.
$\ddagger$~Claude-Opus-5 was written through an agent harness rather than the API
path every other row used (\Cref{tab:main_results}).}
\label{tab:models}
\small
\setlength{\tabcolsep}{1.25pt}
\resizebox{\linewidth}{!}{%
\begin{tabular}{lllrl}
\toprule
Family & Identifier & Short name & Size & Serving \\
\midrule
Anthropic & \texttt{anthropic/claude-haiku-4.5} & Claude-haiku-4.5~\cite{anthropic2025claudehaiku45} & n/d & OpenRouter \\
          & \texttt{anthropic/claude-opus-5}$^\ddagger$ & Claude-opus-5~\cite{anthropic2026claudeopus5} & n/d & agent harness \\
\addlinespace[1pt]
DeepSeek  & \texttt{deepseek/deepseek-chat-v3-0324} & DeepSeek-chat-v3~\cite{deepseekai2024deepseekv3} & 685B/37B & OpenRouter \\
          & \texttt{deepseek/deepseek-r1-0528} & DeepSeek-r1~\cite{deepseekai2025deepseekr1} & 685B/37B & OpenRouter \\
          & \texttt{deepseek/deepseek-v3.2} & DeepSeek-v3.2~\cite{deepseekai2025deepseekv32} & 685B/37B & OpenRouter \\
          & \texttt{deepseek-ai/DeepSeek-V4-Flash} & DeepSeek-V4-Flash~\cite{deepseekai2026deepseekv4highlyefficientmilliontoken} & 284B/13B & NHR@FAU \\
\addlinespace[1pt]
Google    & \texttt{google/gemini-2.5-flash-lite} & Gemini-2.5-flash-lite~\cite{comanici2025gemini25} & n/d & OpenRouter \\
          & \texttt{google/gemini-3.7-flash} & Gemini-3.7-flash~\cite{doshi2026gemini37flash} & n/d & OpenRouter \\
          & \texttt{RedHatAI/gemma-4-31B-it-FP8-block} & Gemma-4-31B-I~\cite{gemmateam2026gemma4} & 31B & NHR@FAU \\
\addlinespace[1pt]
IBM       & \texttt{ibm-granite/granite-4.1-3b} & Granite-4.1-3b~\cite{ibm2026granite41} & 3B & NHR@FAU \\
\addlinespace[1pt]
Meta      & \texttt{meta-llama/llama-3.1-8b-instruct} & Llama-3.1-8b-I~\cite{grattafiori2024llama3} & 8B & OpenRouter \\
          & \texttt{meta-llama/llama-3.3-70b-instruct} & Llama-3.3-70b-I~\cite{grattafiori2024llama3} & 70B & OpenRouter \\
          & \texttt{meta-llama/llama-4-maverick} & Llama-4-maverick~\cite{meta2025llama4} & 400B/17B & OpenRouter \\
\addlinespace[1pt]
Microsoft & \texttt{Microsoft/Phi-4-mini-instruct} & Phi-4-mini-I~\cite{abouelenin2025phi4mini} & 3.8B & NHR@FAU \\
\addlinespace[1pt]
Mistral   & \texttt{GaleneAI/Magistral-Small-2509-FP8-Dynamic} & Magistral-S~\cite{mistralai2025magistral} & 24B & NHR@FAU \\
          & \texttt{mistralai/Ministral-3-14B-Reasoning-2512} & Ministral-3-14B-R~\cite{mistral2025ministral3} & 14B & local vLLM \\
          & \texttt{RedHatAI/Mistral-Small-3.2-24B-Instruct-2506-FP8} & Mistral-S-3.2-24B-I~\cite{mistral2025small32} & 24B & NHR@FAU \\
          & \texttt{mistralai/Mistral-Medium-3.5-128B} & Mistral-M-3.5-128B~\cite{mistral2026medium35} & 128B & NHR@FAU \\
          & \texttt{mistralai/mistral-large-2512} & Mistral-L~\cite{mistral2025large3} & 675B/41B & OpenRouter \\
\addlinespace[1pt]
OpenAI    & \texttt{openai/gpt-4o-mini-2024-07-18} & GPT-4o-mini~\cite{openai2024gpt4omini} & n/d & OpenRouter \\
          & \texttt{openai/gpt-5-nano} & GPT-5-nano~\cite{openai2025gpt5systemcard} & n/d & OpenRouter \\
          & \texttt{openai/gpt-5-mini} & GPT-5-mini~\cite{openai2025gpt5systemcard} & n/d & OpenRouter \\
          & \texttt{openai/gpt-oss-120b} & GPT-oss-120b~\cite{openai2025gptoss} & 120B/5B & NHR@FAU \\
\addlinespace[1pt]
Qwen      & \texttt{qwen/qwen3-32b} & Qwen3-32b~\cite{yang2025qwen3} & 32B & OpenRouter \\
          & \texttt{Qwen/Qwen3.6-35B-A3B-FP8} & Qwen3.6-35B-A3B~\cite{qwen2026qwen36} & 35B/3B & NHR@FAU \\
          & \texttt{qwen3-next-80b-a3b-instruct} & Qwen3-next-80b~\cite{qwen2025qwen3next} & 80B/3B & InnKube \\
          & \texttt{qwen/qwen3.5-122b-a10b} & Qwen3.5-122b~\cite{qwen3.5} & 122B/10B & OpenRouter \\
          & \texttt{qwen/qwen3-235b-a22b-thinking-2507} & Qwen3-235b-T~\cite{yang2025qwen3} & 235B/22B & OpenRouter \\
          & \texttt{qwen/qwen3.5-397b-a17b} & Qwen3.5-397b~\cite{qwen3.5} & 397B/17B & OpenRouter \\
\addlinespace[1pt]
Other     & \texttt{utter-project/EuroLLM-22B-Instruct-2512} & EuroLLM-22B-I~\cite{ramos2026eurollm22btechnicalreport} & 22B & local vLLM \\
          & \texttt{soofi-s-isar-preview} & Soofi-S-Isar~\cite{soofiteam2026sovereignopensourcefoundationmodel} & 32B/3B & InnKube \\
          & \texttt{windprak/open\_steuerllm} & Open\_steuerllm~\cite{Wind_2026} & 28B & local vLLM \\
\bottomrule
\end{tabular}
}
\end{table}

\subsection{Generation Without and With Statute Retrieval}
\label{sec:generation_conditions}

We evaluate two generation conditions over the same case facts: all 32 models
without retrieval, and a subset of seven of them, chosen for cost reasons, with
retrieval; the prompts are reproduced in~\Cref{app:all_prompt}. The
conditions differ in whether statutory text is inserted into the instruction's
context slot and, as a consequence, in the instruction that frames that slot
(\Cref{sec:essay_task}). Except in the oracle condition described below, no
material derived from the reference solution is available during generation.

\paragraph{Generation without retrieval.} Each model receives the fixed German
essay instruction and the facts of one case, and generates a complete legal
opinion (\textit{Gutachten}). No statutory text or other external legal material
is supplied. Every model receives the same 81 cases under this condition; the few essays that came back empty, truncated or errored are reported in~\Cref{sec:paired_results_text}.

\paragraph{Generation with statute retrieval.}
For cost reasons, this condition is run on a subset of seven of the 32 models (\Cref{tab:ladder}). They receive the same case facts with statutory text inserted into the instruction's context slot, retrieved automatically from a corpus of German federal and Bavarian law.
  \Cref{sec:retrieval_setup} describes that corpus, the retrieval pipeline and the arms we vary; \Cref{sec:rag_results} reports what they achieve.

Across both conditions, we set no additional output-token limit, and we do not
set temperature or top-\textit{p}: each request uses the serving endpoint's own
defaults, which are recorded in the run records where the endpoint reports them
(\Cref{sec:release}). If a response contains a
reasoning trace delimited by \texttt{<think>} tags, the trace is removed before
the answer is scored.

\subsection{Statute Corpus and Retrieval Pipeline}
\label{sec:retrieval_setup}

The retrieval task in \gpbam{} has a built-in evaluation set through the norms cited in the case solution.
This lets us gauge retrieval quality independently of the final essay quality, and therefore ablate on the integration and generation aspect separately from the search itself.
That opportunity is why statutory text is the primary context throughout; secondary sources enter as a second knowledge source, evaluated both as context in their own right and as a route to the statutes they cite.

\paragraph{Corpus.}
German public law lies at the intersection of federal regulations and respective state law, with applicability a crucial part of the legal appraisal.
We therefore construct a combined corpus of federal law from the \texttt{gesetze-im-internet} dump maintained by the QuantLaw project~\cite{coupette2021measuring},\footnote{branch \texttt{data}, revision \texttt{67e0567}, a
scrape dated 2026-02-09} parsed into 106{,}354 provisions across 5{,}936 law books, and of Bavarian state law collected from \textit{BAYERN.RECHT}, the portal of the Bayerische Staatskanzlei~\cite{bayernrecht}, comprising all 7{,}008 available provisions from 720 statutes.
The combined index holds 113{,}362 provisions.
Both halves are built the same way: we extract the provision text, chunk at law-paragraph borders, and fold the law metadata (name, law book, paragraph) into the indexed text.
The federal dump contains no \emph{Landesrecht} at all, and 38.7\% of the distinct norms a reference solution cites are exactly that,
so a federal-only corpus caps recall at 54.2\% against the combined corpus's 91.3\% (\Cref{tab:reference_coverage}).

As an additional knowledge source we index \emph{Leitsätze}, headnotes produced over legal decisions by a German legal publisher as secondary literature.
Headnotes are short - 306 characters and 40 words on the mean, 285 characters at the median - but they carry explicit guidance on how the law was applied and summarise the decision at the same time.
We index 38{,}966 of them, extracted from the German legal decision corpus of Urchs et al.~\cite{urchs2021gdc} and covering the Bavarian public-law courts over 2015--2020.
Each carries the \emph{Normenkette} of its decision, which is what makes a second hop to statutes possible.
Roughly 4\% are \emph{amtliche} headnotes written by the court itself, the remainder editorial work by a legal publisher.

\paragraph{Query construction.}
Before generation, Qwen3.6-35B-A3B rewrites each case into three to five stand-alone legal concepts or questions, an implicit query diversification scheme.
This moves some of the legal knowledge boundary onto the rewriter, but both previous retrieval studies~\cite{ma-etal-2023-query,gao-etal-2023-precise} and our own find that it clearly outperforms other rewriting strategies as well as the raw task description (tripling gold-norm recall@$50$ against the raw case text). The concept rewrites are generated once and reused across every arm that searches, so that two search arms differ in the retrieval component under study and not in the query.

\paragraph{Search and fusion.}
For dense retrieval, we use LanceDB's Sentence Transformers~\cite{reimers-2019-sentence-bert}
integration with \texttt{jina-embeddings-v5-text-small-\allowbreak retrieval},
a task-targeted distillation of the Jina v5 text family~\cite{akram2026jinav5}.
We retain the adapter's default encoding behaviour for both corpus indexing and
query embedding and combine this dense channel with lexical BM25~\cite{robertson2009bm25} retrieval over
the same corpus.
For each rewritten query, reciprocal rank fusion (RRF; $k_{\mathrm{RRF}}{=}60$)~\cite{cormack2009rrf} first merges the dense and
BM25 rankings.
RRF is then applied a second time across the resulting rankings from all
rewritten queries.
The ten highest-ranked passages are passed to the generator unless a different
passage budget is specified.

In short, the two retrieval methods produce candidate rankings for each rewritten query, and RRF is applied twice: first to merge the dense and BM25 rankings within each query, and then to merge the results across the rewritten queries before the highest-ranked passages are passed to the generator.

\paragraph{Retrieval arms}
The retrieval arms vary passage budget and components (\Cref{tab:ladder}).
The \texttt{cite} arms resolve norms named by a separate rewriter;
\texttt{cite-only} provides them alone, while \texttt{cite+RRF} fills the remaining budget with hybrid search.
Other arms re-rank with
Qwen3-Reranker-4B~\cite{zhang2025qwen3embedding}, resolve the \emph{Normenketten} of twenty retrieved headnotes,
or supply headnotes directly.
Finally, an \emph{oracle} supplies the full text of provisions cited by the solution.
It bounds corpus retrieval and is not deployable.

\subsection{LLM Judge Ensemble}
\label{sec:judges}

Complete legal case solutions are open-ended: legally adequate answers may differ substantially in wording while expressing equivalent legal analyses.
Consequently, lexical or shallow semantic similarity to a reference solution cannot by itself assess issue spotting, doctrinal assessment, completeness, structure, or subsumption.
At benchmark scale, manual assessment of such responses requires costly annotation by legal experts~\cite{10.1145/3769126.3769227, chlapanis-etal-2025-greekbarbench}.
Recent legal benchmarks therefore employ rubric- or reference-guided LLM judges as scalable semantic evaluators. Comparisons with expert annotation show that carefully specified judges \textit{can} provide useful assessments~\cite{fan2026lexambenchmarkinglegalreasoning, chlapanis-etal-2025-greekbarbench,nagl2026bengerbenchmarkingllmsystems}.
These studies also identify differences in judge severity, prompt sensitivity, and model-family preference~\cite{GU2026101253,chlapanis-etal-2025-greekbarbench,nagl2026bengerbenchmarkingllmsystems, panickssery2024llmevaluatorsrecognizefavor}.

Each essay is therefore graded independently by three open-weight LLM judges drawn from three different developers: GPT-oss-120b,
Qwen3.6-35B-A3B, and DeepSeek-V4-Flash. Panels of this kind have been reported to grade legal answers more reliably than their strongest single member:
on German examination scripts, as shown by~\cite{zubaer2026gradelegalautomatedgradinggerman}, pooling three models raises agreement with expert marks by up to $0.15$ QWK over the best member when the full grading context is supplied,
and an ensemble judge
is also what LEXam validates against expert
assessment~\cite{fan2026lexambenchmarkinglegalreasoning}. The composition
additionally reduces dependence on any single evaluator and limits the influence
of developer or model-specific evaluation preferences.

Because judges may nevertheless differ systematically in severity,
we aggregate their essay-level scores using the median. With three judges, the median
retains the central assessment while limiting the influence of one atypically high
or low score. We report absolute and rank agreement between the seats at essay and model
level in~\Cref{sec:judge_analysis}.

All judges receive the same evidence in the prompt: the case, its reference solution, and the generated essay.
The instructions guide each judge to align the reference and generated answers by section and assess correctness, completeness, legal reasoning, structure, depth of subsumption, and statutory citations, scoring how far the essay matches the solution's results, subsumption, and citations, and crediting deviations from the solution only if they are legally sound.
Each judge is instructed to return a score from 0.0 to 1.0 in increments of 0.1.
The complete judge prompt is reproduced in~\Cref{appendix:judge_prompt}.
To anchor the judge scale, we additionally submit three controls for each case: its reference solution, another case's reference solution, and an answer that repeats the case facts (\Cref{sec:scale_check}).

\subsection{Evaluation Metrics}
\label{sec:metrics}

\paragraph{Ensemble Judge Score} Let $s_{mij}\in[0,1]$ be the score judge $j$ assigns to the essay that model $m$ writes for case $i$, and let $\mathcal{J}_{mi}$ be the set of judges that returned a score for that essay. The per-essay score and a model's reported \textbf{Score} are
\begin{align}
  s_{mi} &= 100 \cdot \operatorname*{median}_{j \in \mathcal{J}_{mi}} s_{mij},
  \label{eq:essay_score}\\
  S_m    &= \frac{1}{81} \sum_{i=1}^{81} s_{mi}.
  \label{eq:model_score}
\end{align}
Missing judge scores are skipped rather than imputed, so $\mathcal{J}_{mi}$ may contain fewer than three judges.
The error term reported with $S_m$ in ~\Cref{tab:main_results} is its
standard error, $\hat{\sigma}_m/\sqrt{81}$, where $\hat{\sigma}_m$ is the sample
standard deviation of the 81 case-level scores $s_{mi}$.

\paragraph{Self-preference adjustment}
Three judges are also generators and eleven more share vendors.
\textbf{Deb}iased in \Cref{tab:main_results} subtracts the excess score that judges give to related models before taking the median.
We estimate that excess at matched essay quality: each judge's scores become percentiles within that judge, we fit them against the mean percentile of the other two on the same essay, and read the offset off that fit.

\paragraph{Legal Reference Similarity} For each generated essay and its reference solution, we extract the unique pairs of law book and provision, resolving both \S- and Art.-form citations.\footnote{Our extractor, \texttt{refs.py}: \url{https://github.com/wendli01/gpbam/blob/fb7796e/experiments/analysis/refs.py}.} Legal reference similarity is the Jaccard overlap of these pairs, $|A_{mi}\cap R_i|/|A_{mi}\cup R_i|$, where $A_{mi}$ and $R_i$ are the pairs extracted from the generated essay and the reference solution, respectively. We scale essay-level values by 100. The widely used \texttt{refex}\footnote{\url{https://github.com/openlegaldata/legal-reference-extraction}} resolves only \S-form citations and so misses constitutional and Bavarian provisions, finding about 21 norms per solution instead of 52.

\paragraph{Article Recitation} For each provision in the \emph{GPBam laws} and \emph{Most-cited laws} sets
defined in \Cref{sec:article_recitation_task}, the generated text is compared
with its stored target using ROUGE-L F1 without stemming. This metric measures
longest-common-subsequence overlap with the statutory wording. For each model
and provision set, we scale scores by 100 and report the mean over the set's 100
provisions, or over the provisions actually scored for the four models noted in~\Cref{gpbam:para:scope}).

\paragraph{Retrieval Recall} Gold-norm recall@$k$ is the share of the statutory references a reference solution cites that appear among the $k$ passages an arm returns for that case, pooled over the 81 cases rather than averaged per case, so a solution weighs in proportion to how much law it names; $k$ is the arm's own passage budget, ten unless marked $k{=}50$.
State-law recall@$k$ is the same share over the \emph{Landesrecht} references alone, and cases with a hit@$k$ the share of cases where at least one cited norm is returned.

In short, gold-norm recall@$k$ asks what share of all cited norms across the 81 cases is retrieved within the passage budget, state-law recall@$k$ asks the same question for \emph{Landesrecht} only, and hit@$k$ asks in what share of cases at least one cited norm is found.

\paragraph{Across Model Relationships}
To examine which model-level characteristics are associated with essay quality,
we compute Spearman rank correlations between each model's mean ensemble score
and its Article Recitation score on either provision set, its Artificial
Analysis index, its answer tokens, its generation cost,
and its Legal Reference Similarity (Table~\Cref{tab:essay_corr_combined}). We
report the \norag{} condition only.
Each model is one observation, and only the model-level point estimates are
correlated. Percentile 95\% confidence intervals use 10,000 bootstrap resamples
of the paired model observations; pairwise-complete data are used, so $n$ is
smaller for the Artificial Analysis index, which does not list every model we
evaluate, and for generation cost, where provider pricing is unavailable.

\paragraph{Inter-judge agreement} We compare every judge pair using Spearman's $\rho$ and quadratic-weighted kappa (QWK, $\kappa_w$). At model level, one observation is a generation model and each judge's value is its mean over that model's available essays; percentile 95\% confidence intervals are computed from 10,000 bootstrap resamples, each formed by sampling the 32 generation models with replacement. At essay level, one observation is an essay scored by both judges, and confidence intervals use 10,000 cluster-bootstrap resamples of the 32 generation-model clusters with replacement, keeping all essays from each selected model together to preserve dependence among its 81 essays. Spearman measures rank agreement, whereas QWK measures agreement in absolute score levels: at essay level the $0$--$1$ scores are placed on the judges' own fixed 11-point ordinal grid ($0.0$--$1.0$ in steps of $0.1$, rescaled to $0$--$100$), pinned to all eleven levels rather than to those a given pair happens to use, so that the index spacing, and hence the quadratic weights, means the same thing for every pair; the few judgements that land off that grid are parser strays and are snapped to the nearest step. At model level, where the values are means rather than grid points, we use the bin-width-to-zero limit of the same statistic. For each judge pair, we compute the coefficients using only essays or models scored by both judges. See~\Cref{tab:judge_agreement_norag}

\section{Results}
\label{sec:results}

\subsection{Performance on Complete Legal Opinions}
\label{sec:paired_results_text}

\paragraph{Overall performance.} \Cref{tab:main_results} shows that \gpbam{} is not saturated by the current model population.
Without retrieval, the 32 models occupy $4.81$--$76.42$ points of the $0$--$100$
scale, and the distribution is concentrated in its lower half.
The median model score is $30.2$, 21 of the 32 fall below $33$, and nine remain under $20$.
However, two models are clearly separated from the remainder. Claude-opus-5 ($76.42$) and Gemini-3.7-flash ($67.10$) precede a $23.3$-point discontinuity to the strongest open-weight entry, DeepSeek-V4-Flash ($43.77$).
This single interval exceeds the entire range from rank three to rank
eighteen ($43.77$ to $26.36$), which comprises sixteen models.
Below the leading two, the table constitutes a dense band rather than an ordering.
The distinctions the measurements support are those between Claude-opus-5, Gemini-3.7-flash, and the remaining thirty.
The two recitation
columns (\textit{Art. recit.}) and the external index (\textit{AA}) are discussed in \Cref{sec:predictors}.

Non-response is negligible except at the bottom of the table.
Over the $2{,}592$ essays (32 models over 81 cases), twenty-one ($0.8\%$) came back empty, truncated or errored, twelve of them Phi-4-mini-I ($15\%$ of its cases).%

\begin{table}[htbp]
\centering
\caption{\small \textbf{Model performance without retrieval, for both \gpbam{} tasks.} \textbf{Median} is the mean of the per-case median over three judges, see per-judge columns, the subscript is its standard
error, $\mathrm{sd}/\sqrt{81}$
\textbf{Deb.} removes self-preference of a judge related to the model: M same model, V same vendor, blank if none.
  Cells are colored  (red below and blue above, saturating at $\pm3$\,SD) for the deviation from the median of that row; \textbf{best in class}, \underline{second best} and \textit{best open-weights} results are marked.
  \textbf{Ref.} is the Reference Similarity  between references cited in the essay and in the solution.
  \textbf{GP} and \textbf{Cited} are ROUGE-L F1 for reciting the most frequent provisions in \gpbam{} solutions and German court decisions respectively. $^{\ddagger}$Scored on the 73 \emph{\gpbam{} laws} provisions only, which may differ slightly from a full-set score; see \Cref{sec:article_recitation_task}.
  \textbf{AA} is \textbf{Gen}eral capability via Artificial Analysis index, where available.
  $\dagger$~run through an agent harness, not API, so it carries a harness system prompt where every other row has none; transcripts show no access to anything but the prompt file.
  }
\scriptsize
\setlength{\tabcolsep}{1.25pt}
\renewcommand{\arraystretch}{.94}
\resizebox{\linewidth}{!}{%
\begin{tabular}{l @{\hspace{1em}} r r @{\hspace{0.5em}} r r r @{\hspace{0.5em}} r @{\hspace{1em}} r r @{\hspace{1em}} r}
\toprule
 & \multicolumn{6}{c}{Essay writing} & \multicolumn{2}{c}{Art. recit.} & \multicolumn{1}{c}{Gen.} \\
\cmidrule(lr){2-7} \cmidrule(lr){8-9} \cmidrule(lr){10-10}
 & \multicolumn{2}{c}{Judge panel} & \multicolumn{3}{c}{Per judge} & & & & \\
\cmidrule(lr){2-3} \cmidrule(lr){4-6}
Model & Median & Deb. & GPT-oss & Qwen3.6 & DS-V4 & Ref. & GP & Cited & AA \\
\midrule
Claude-opus-5$^{\dagger}$ & \textbf{76.42}$_{\pm2.35}$ &  & \cellcolor[HTML]{FCD3BD}70.12 & \cellcolor[HTML]{154F8C}\textcolor{white}{90.49} & \cellcolor[HTML]{F7F7F7}74.32 & \textbf{39.63} & \underline{74.54} & \textbf{89.97} & \textbf{63} \\
Gemini-3.7-flash& \underline{67.10}$_{\pm1.92}$ &  & \cellcolor[HTML]{F7F7F7}62.35 & \cellcolor[HTML]{114882}\textcolor{white}{79.01} & \cellcolor[HTML]{F9EEE8}61.17 & \underline{36.53} & \textbf{74.92} & \underline{69.83} & \underline{56} \\
DeepSeek-V4-Flash & \textit{43.77}$_{\pm1.58}$ & \cellcolor[HTML]{A01228}\textcolor{white}{41.05$^{\mathrm{M}}$} & \cellcolor[HTML]{F7F7F7}41.54 & \cellcolor[HTML]{3783BB}53.83 & \cellcolor[HTML]{FDD9C5}37.78 & \textit{22.21} & 50.82 & 39.06 & \textit{52} \\
Qwen3.5-397b & 41.48$_{\pm1.49}$ & \cellcolor[HTML]{FDDCC9}40.86$^{\mathrm{V}}$ & \cellcolor[HTML]{F7F7F7}41.30 & \cellcolor[HTML]{6CADD1}50.37 & \cellcolor[HTML]{F5AA88}34.32 & 20.69 & 40.74 & 34.38 & 34 \\
Mistral-L & 37.84$_{\pm1.37}$ &  & \cellcolor[HTML]{F7F7F7}38.70 & \cellcolor[HTML]{89C0DB}46.42 & \cellcolor[HTML]{F2A07E}31.11 & 19.86 & \textit{56.46} & \textit{50.54} & 16 \\
Mistral-M-3.5-128B & 37.41$_{\pm1.39}$ &  & \cellcolor[HTML]{F7F7F7}38.77 & \cellcolor[HTML]{A8D0E4}44.81 & \cellcolor[HTML]{E7876B}29.88 & 18.47 & 36.09 & 28.84 & 30 \\
DeepSeek-v3.2 & 36.79$_{\pm1.19}$ & \cellcolor[HTML]{F8F2EE}36.67$^{\mathrm{V}}$ & \cellcolor[HTML]{F7F7F7}37.04 & \cellcolor[HTML]{8DC2DC}44.57 & \cellcolor[HTML]{F3A381}29.63 & 17.29 & 33.84 & 27.29 & 33 \\
Gemini-2.5-flash-lite & 35.19$_{\pm1.19}$ &  & \cellcolor[HTML]{F7F7F7}33.83 & \cellcolor[HTML]{80BAD8}41.98 & \cellcolor[HTML]{FDD8C4}30.00 & 14.80 & 21.09 & 20.28 & 11 \\
GPT-5-mini & 34.81$_{\pm1.17}$ & \cellcolor[HTML]{F7F7F7}34.81$^{\mathrm{V}}$ & \cellcolor[HTML]{F7F7F7}35.80 & \cellcolor[HTML]{BDDBEA}40.62 & \cellcolor[HTML]{F5AD8C}29.01 & 12.33 & 21.74 & 17.41 & 26 \\
Claude-haiku-4.5 & 34.81$_{\pm1.08}$ &  & \cellcolor[HTML]{F7F7F7}36.30 & \cellcolor[HTML]{D6E7F1}39.51 & \cellcolor[HTML]{E17860}26.60 & 13.59 & 25.44 & 16.07 & 30 \\
DeepSeek-r1 & 33.21$_{\pm1.19}$ & \cellcolor[HTML]{F9ECE4}32.96$^{\mathrm{V}}$ & \cellcolor[HTML]{F7F7F7}33.46 & \cellcolor[HTML]{A4CEE3}39.75 & \cellcolor[HTML]{FAC6AD}28.40 & 17.21 & 27.25 & 22.42 & 20 \\
Qwen3.5-122b & 32.96$_{\pm1.18}$ & \cellcolor[HTML]{F7F7F7}32.96$^{\mathrm{V}}$ & \cellcolor[HTML]{F7F7F7}33.58 & \cellcolor[HTML]{90C4DD}40.99 & \cellcolor[HTML]{F6B292}27.12 & 15.79 & 31.44 & 25.50 & 33 \\
Gemma-4-31B-I & 32.90$_{\pm1.07}$ &  & \cellcolor[HTML]{F7F7F7}32.62 & \cellcolor[HTML]{6EAED2}41.60 & \cellcolor[HTML]{FCE0D0}29.63 & 10.72 & 18.53 & 14.30 & 30 \\
Qwen3-235b-T & 31.11$_{\pm1.10}$ & \cellcolor[HTML]{F7F7F7}31.11$^{\mathrm{V}}$ & \cellcolor[HTML]{F7F7F7}32.72 & \cellcolor[HTML]{CEE4EF}36.54 & \cellcolor[HTML]{EA8E71}24.20 & 12.09 & 20.05 & 16.15 & 20 \\
Qwen3.6-35B-A3B & 30.62$_{\pm1.00}$ & \cellcolor[HTML]{BE3036}\textcolor{white}{28.27$^{\mathrm{M}}$} & \cellcolor[HTML]{F7F7F7}30.37 & \cellcolor[HTML]{9FCCE2}36.91 & \cellcolor[HTML]{F5AD8C}23.58 & 14.56 & 23.56 & 20.27 & 32 \\
Qwen3-next-80b & 30.62$_{\pm0.94}$ & \cellcolor[HTML]{F7F7F7}30.62$^{\mathrm{V}}$ & \cellcolor[HTML]{F7F7F7}33.95 & \cellcolor[HTML]{EEF3F5}34.81 & \cellcolor[HTML]{DD7059}23.83 & 12.57 & 14.27 & 13.60 & 14 \\
GPT-oss-120b & 29.75$_{\pm0.99}$ & \cellcolor[HTML]{E58267}28.15$^{\mathrm{M}}$ & \cellcolor[HTML]{B0D4E7}36.30 & \cellcolor[HTML]{F7F7F7}30.74 & \cellcolor[HTML]{E48066}21.48 & 12.19 & 18.36 & 13.75 & 24 \\
DeepSeek-chat-v3 & 26.36$_{\pm0.82}$ & \cellcolor[HTML]{F7F7F7}26.36$^{\mathrm{V}}$ & \cellcolor[HTML]{F7F7F7}25.86 & \cellcolor[HTML]{B1D5E7}31.36 & \cellcolor[HTML]{FCD4BE}21.73 & 13.64 & 25.00 & 24.06 & 15 \\
Soofi-S-Isar & 26.11$_{\pm0.91}$ &  & \cellcolor[HTML]{F7F7F7}26.48 & \cellcolor[HTML]{9ECBE1}33.09 & \cellcolor[HTML]{FCD4BE}22.35 & 14.62 & 26.16 & 24.92 & -- \\
Llama-4-maverick & 24.44$_{\pm0.68}$ &  & \cellcolor[HTML]{F7F7F7}25.56 & \cellcolor[HTML]{E5EEF4}27.28 & \cellcolor[HTML]{FCDDCB}22.22 & 10.96 & 23.11 & 36.13 & 14 \\
Magistral-S & 23.33$_{\pm0.70}$ &  & \cellcolor[HTML]{F7F7F7}23.09 & \cellcolor[HTML]{D0E5F0}26.79 & \cellcolor[HTML]{FBE3D5}20.49 & 10.33 & 14.06 & 14.85 & 11 \\
Mistral-S-3.2-24B-I & 21.98$_{\pm0.62}$ &  & \cellcolor[HTML]{F7F7F7}21.85 & \cellcolor[HTML]{DBEAF2}24.57 & \cellcolor[HTML]{FAE9DF}20.00 & 9.79 & 15.49 & 15.76 & 11 \\
GPT-5-nano & 21.48$_{\pm0.66}$ & \cellcolor[HTML]{F7F7F7}21.48$^{\mathrm{V}}$ & \cellcolor[HTML]{F7F7F7}22.28 & \cellcolor[HTML]{EFF3F5}23.09 & \cellcolor[HTML]{FCDECC}19.01 & 10.08 & 14.06 & 11.77 & 20 \\
GPT-4o-mini & 19.51$_{\pm0.66}$ & \cellcolor[HTML]{F7F7F7}19.51$^{\mathrm{V}}$ & \cellcolor[HTML]{F7F7F7}18.27 & \cellcolor[HTML]{BFDCEB}22.96 & \cellcolor[HTML]{F8F1ED}17.53 & 6.75 & 16.53 & 16.30 & 7 \\
Open\_steuerllm & 17.78$_{\pm0.63}$ &  & \cellcolor[HTML]{F7F7F7}16.91 & \cellcolor[HTML]{D3E6F0}20.37 & \cellcolor[HTML]{F8F1ED}16.17 & 7.27 & 14.01$^{\ddagger}$ & 17.16 & -- \\
Llama-3.3-70b-I & 16.79$_{\pm0.63}$ &  & \cellcolor[HTML]{F7F7F7}15.56 & \cellcolor[HTML]{E2EDF3}17.53 & \cellcolor[HTML]{F7F7F7}15.56 & 6.22 & 14.52 & 17.12 & 9 \\
Qwen3-32b & 13.21$_{\pm0.82}$ & \cellcolor[HTML]{F7F7F7}13.21$^{\mathrm{V}}$ & \cellcolor[HTML]{EEF3F5}13.70 & \cellcolor[HTML]{F8F3F1}12.35 & \cellcolor[HTML]{F7F7F7}12.84 & 7.62 & 10.26 & 9.95 & 11 \\
EuroLLM-22B-I & 9.26$_{\pm0.76}$ &  & \cellcolor[HTML]{F9ECE4}7.90 & \cellcolor[HTML]{E9F0F4}10.74 & \cellcolor[HTML]{F7F7F7}9.38 & 4.17 & 9.86$^{\ddagger}$ & 12.11 & -- \\
Ministral-3-14B-R & 8.77$_{\pm0.75}$ &  & \cellcolor[HTML]{F7F7F7}9.26 & \cellcolor[HTML]{FCE1D2}6.42 & \cellcolor[HTML]{E5EEF4}10.99 & 8.66 & 9.28$^{\ddagger}$ & 12.30 & 10 \\
Granite-4.1-3b & 7.53$_{\pm0.57}$ &  & \cellcolor[HTML]{F9ECE4}5.80 & \cellcolor[HTML]{F7F7F7}7.28 & \cellcolor[HTML]{E4EEF3}9.14 & 3.83 & 7.80 & 9.73 & 4 \\
Phi-4-mini-I & 5.43$_{\pm0.58}$ &  & \cellcolor[HTML]{F9EEE8}4.44 & \cellcolor[HTML]{F7F7F7}5.56 & \cellcolor[HTML]{E1EDF3}7.65 & 3.84 & 7.68 & 8.46 & 6 \\
Llama-3.1-8b-I & 4.81$_{\pm0.68}$ &  & \cellcolor[HTML]{FAE9DF}3.46 & \cellcolor[HTML]{F7F7F7}5.31 & \cellcolor[HTML]{DFEBF3}7.65 & 3.07 & 7.42 & 10.20 & 7 \\
\bottomrule
\end{tabular}
}

  \label{tab:main_results}
\end{table}
\paragraph{What the scale means.}
\label{sec:scale_check}

Three synthetic answers per case, judged by the same three seats under the same instruction, anchor the $0$--$100$ axis.
A case's reference solution, submitted as an answer, scores $100.0$ on all 81 cases, so the top of the scale is attainable rather than notional.
A \emph{different} case's reference solution - expert-written, correctly structured, statute-dense German legal prose answering the wrong question - scores $0.0$, as does echoing the \emph{Sachverhalt} back.
The panel median is invariant at both anchors; individual seats depart only marginally.
The anchors fix the endpoints, and those are the easy case - the judge is given the reference solution, so spotting a wrong-case answer needs no legal judgement.
They show the scale is not degenerate, not that it is linear between; \Cref{sec:opus_probe} locates the remaining disagreement in the upper interior, where the field's spread lives.

\paragraph{Reference Similarity}

Legal Reference Similarity (\textit{Ref.}) gives another measure of essay quality, with a similar gap: the leading two models
achieve an overlap of $39.63\%$ and $36.53\%$ with the solution references, while the third model falls below one quarter, and the final nine rows remain under $9\%$.
This column tracks the essay-score ranking closely at $\rho{=}0.94$ with only three models exhibiting a difference of more than five positions.
Splitting the Jaccard similarity shows that mean precision is $0.56$ but mean recall only $0.16$: a model cites $14.5$ provisions where the solution cites $52$.
These results indicate that the dominant citation failure is incomplete coverage, not lacking precision over a small citation set.

\paragraph{Scale and family.} We observe that neither parameter count nor developer family accounts for the ordering of the models.
Gemma-4-31B-I ($32.90$) ranks above DeepSeek-chat-v3 ($26.36$) and Llama-4-maverick ($24.44$), both 400B or larger in total parameters compared to 31B, while Qwen3-32b attains $13.21$ at comparable scale. Variation within a single family matches variation across families with Mistral $8.77$--$37.84$, Qwen $13.21$--$41.48$.
Building for a language or a legal domain is no substitute either, and the three systems selected outside the general-purpose developer families place 19th, 25th and 28th.
Soofi-S-Isar, pretrained on roughly 27 trillion tokens with German deliberately up-weighted~\cite{soofiteam2026sovereignopensourcefoundationmodel}, reaches $26.11$ and leads the other two, but stays below Gemma-4-31B-I ($32.90$) at comparable total scale and below the other two A3B models, Qwen3.6-35B-A3B and Qwen3-next-80b ($30.62$ each); EuroLLM-22B, built for the official EU languages~\cite{ramos2026eurollm22btechnicalreport}, scores $9.26$.
Open\_steuerllm, adapted to German tax law~\cite{Wind_2026}, ranks 25th at $17.78$ points, below three of the four general-purpose models within a few billion parameters of its size:\footnote{Open\_steuerllm is 28B and scores $17.78$. The four are Gemma-4-31B-I (31B, $32.90$), Magistral-S (24B, $23.33$), Mistral-S-3.2-24B-I (24B, $21.98$) and Qwen3-32b (32B, $13.21$); it outscores only the last. Sizes from \Cref{tab:models}, scores from \Cref{tab:main_results}.} adaptation to one legal domain shows no advantage for public-law examination writing here.

\paragraph{Judge behaviour.}

The three seats differ in level more than in ordering: Qwen3.6-35B-A3B returns the highest of the three scores on 26 of the 32 rows and DeepSeek-V4-Flash the lowest on 25, but the between-judge spread reaches  $20.4$ points on Claude-opus-5 and at most $4.6$ points on the four weakest models, where it also reverses direction, the otherwise strictest judge becoming the most lenient. The median in~\Cref{eq:essay_score} absorbs this.
Self-preference is narrower: the only corrections above one point fall on the three models that are themselves judges ($-2.72$, $-2.35$, $-1.60$), while the eleven vendor-related rows move by at most $0.62$ points.
Model order is unchanged, but the gap between DeepSeek-V4-Flash and Qwen3.5-397b narrows (2.29 to 0.19).

\subsection{What Predicts Essay Quality}
\label{sec:predictors}

\begin{figure}[htbp]
  \centering
  \begin{subfigure}[t]{0.49\textwidth}
    \centering
    \includegraphics[width=\linewidth]{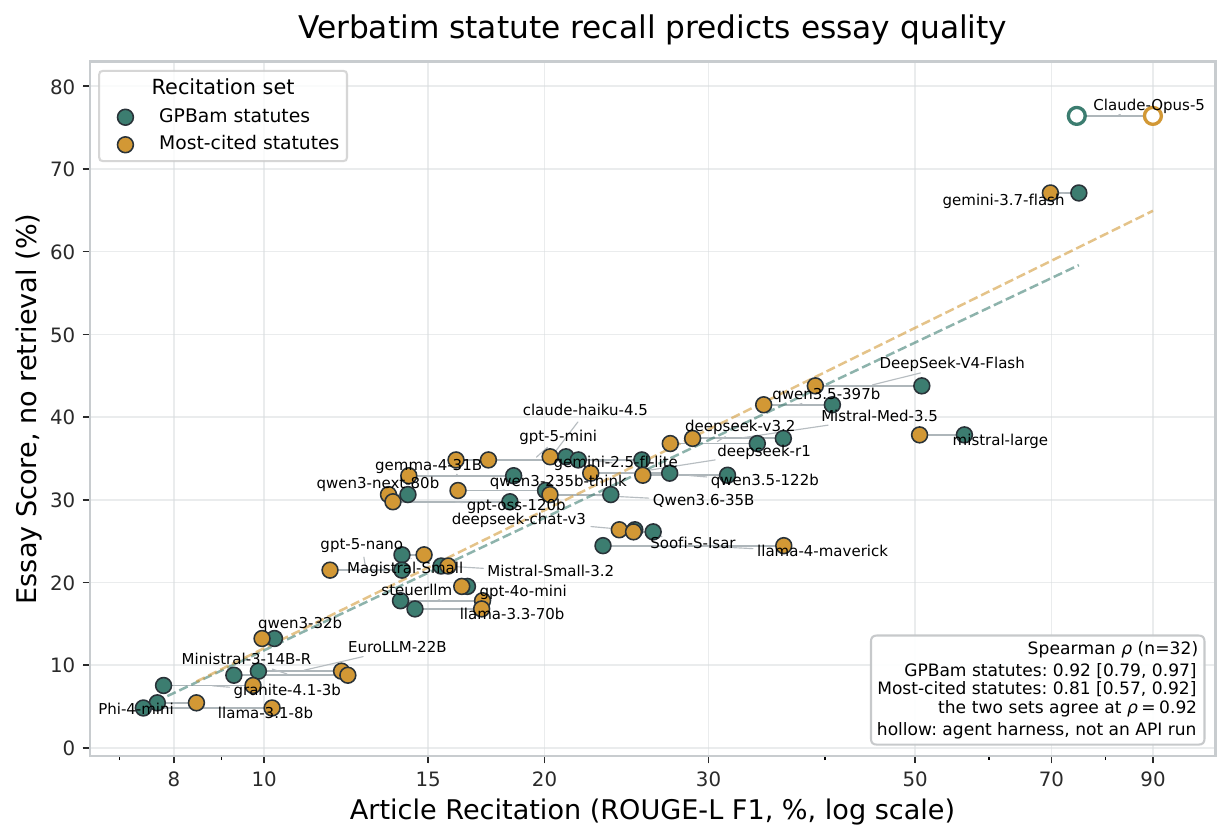}
\caption{Article Recitation on both provision sets
  ($\rho{=}0.92$ between sets).}
    \label{fig:recitation_vs_essay}
  \end{subfigure}
  \hfill
  \begin{subfigure}[t]{0.49\textwidth}
    \centering
    \includegraphics[width=\linewidth]{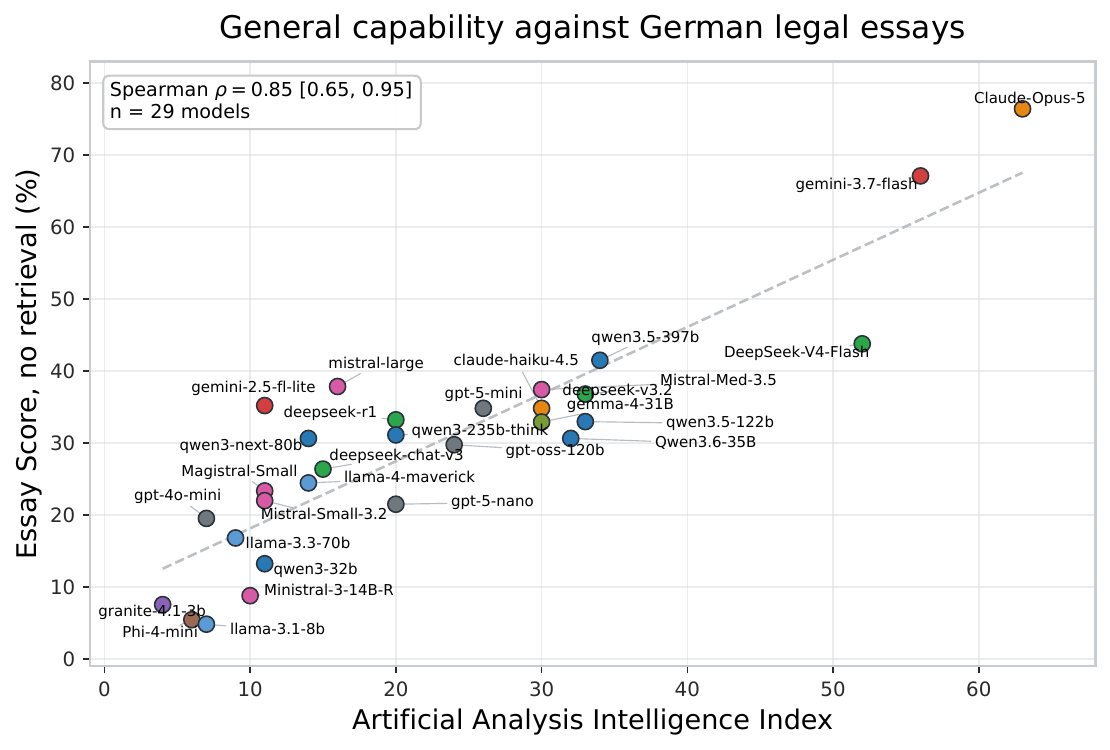}

\caption{Artificial Analysis Intelligence Index.}
    \label{fig:aa_index_vs_essay}
  \end{subfigure}

\caption[External predictors of \norag{} essay quality.]{\textbf{External predictors of \norag{} essay quality.}
Each point represents one model. Essay scores averaged across 81 cases.
Panels report Spearman $\rho$ with 95\% percentile-bootstrap CIs.
AA Intelligence Index values were taken from the public Artificial Analysis
model pages (index v4.1.1, accessed 2026-08-17). For 11 of the 29 models,
Artificial Analysis reports the value as an estimate pending independent
evaluation\protect\footnotemark{}; we use these values as published.
For Claude-Opus-5 and DeepSeek-V4-Flash, only a max-effort configuration is
listed, whereas our runs use high effort.
Later index versions rescale the values.}
  \label{fig:knowledge_vs_essay}
\end{figure}

Article Recitation measures both general and domain-specific statutory knowledge while
the Artificial Analysis Index~\cite{artificial_analysis_2026}
measures general capability. Essay score correlates with recitation at
$\rho{=}0.92$ {\tiny$[0.79,\,0.98]$} on \emph{\gpbam{} laws} and $0.81$
{\tiny$[0.56,\,0.92]$} on \emph{Most-cited laws}, and with the index at $0.85$
{\tiny$[0.65,\,0.95]$} over its 29 rated models.
These external-predictor relationships are visualized in
\Cref{fig:knowledge_vs_essay}, while
\Cref{tab:essay_corr_combined} summarizes all model-level associations.

The Artificial Analysis index correlates with essay score at $\rho{=}0.85$ (\Cref{fig:aa_index_vs_essay}), so \gpbam{} ranks the model field much as the general capability AA index already does. However, \gpbam{} adds further insight where the two disagree.
Four models carry the same index value of 11 and still span $13.21$ to $35.19$ essay points, which the index cannot resolve.
Mistral-L is the clearest case: an index of 16 puts it in the lower half of the rated field, yet it scores $37.84$, ahead of eleven models the index rates at 20 or above, and it holds the best recitation scores of any open-weight system in the study.

After controlling for the AA index, recitation still predicts essay score ($\rho{=}0.75$; $0.64$ on \emph{Most-cited laws}).
Length also correlates at $0.71$.
Reference Similarity is strongest at $\rho{=}0.94$
{\tiny$[0.84,\,0.98]$}, but is extracted from the graded essays and not
independent evidence. These associations don't answer whether providing missing norms improves essays, but our RAG experiments do.

\footnotetext{Gemini-3.7-flash, DeepSeek-v3.2, Gemini-2.5-flash-lite,
DeepSeek-r1, Qwen3-next-80b, GPT-5-nano, GPT-4o-mini, Llama-3.3-70b-I,
Ministral-3-14B-R, Granite-4.1-3b, Llama-3.1-8b-I.}

\begin{table}[htbp]
\centering

\caption{\textbf{Associations with essay quality without retrieval.}
Spearman correlations between mean essay score and model-level metrics across up
to 32 models; brackets show 95\% bootstrap CIs. Generation costs for self-hosted
models use OpenRouter list prices.}
\label{tab:essay_corr_combined}
\small
\setlength{\tabcolsep}{1.5pt}
\begin{tabular}{l r r@{\hspace{3pt}}l}
\toprule
Predictor & $n$ & \multicolumn{2}{c}{$\rho$ {\tiny[95\% CI]}} \\
\midrule
\multicolumn{4}{l}{\emph{Knowledge}} \\
Article Recitation, GPBam & 32 & 0.92 & {\tiny$[0.79,\,0.98]$} \\
Article Recitation, Most cited & 32 & 0.81 & {\tiny$[0.56,\,0.92]$} \\
Artificial Analysis index & 29 & 0.85 & {\tiny$[0.65,\,0.95]$} \\
\addlinespace[2pt]
\multicolumn{4}{l}{\emph{Length}} \\
Answer Tokens (k) & 32 & 0.71 & {\tiny$[0.51,\,0.83]$} \\
\addlinespace[2pt]
\multicolumn{4}{l}{\emph{Cost}} \\
Gen. Cost (\$) & 27 & 0.77 & {\tiny$[0.52,\,0.90]$} \\
\addlinespace[2pt]
\multicolumn{4}{l}{\emph{Same essay}} \\
Legal Ref. Sim. & 32 & 0.94 & {\tiny$[0.84,\,0.98]$} \\
\bottomrule
\end{tabular}
\end{table}

\subsection{Can Retrieval Close the Knowledge Gap?}
\label{sec:rag_results}

The following experiment tests whether this knowledge gap is indeed a binding constraint.
We observe that the correlation between judged essay quality and reference similarity is high, which highlights the importance of correct context sourcing.
As article recitation also closely tracks essay quality, we attempt to fill the legal knowledge gap via retrieval over a law corpus (\Cref{sec:retrieval_setup}).
By providing a set of relevant norms and their text, we attempt to equalize across different levels of inherent static knowledge and ablate on the dynamic ability to subsume and apply a law.

The corpus is not the binding constraint.
A federal-only index caps recall at 54.2\% of the gold norms, against 91.3\% for the combined corpus (\Cref{tab:reference_coverage}).

\begin{table}[htbp]
\centering
\caption{\textbf{What the reference solutions cite, and how much of it the retrieval corpus holds.} Every distinct statutory reference in the 81 reference solutions, counted once per case, (\textcolor{black}{ extracted with the dual \S/Art.\ extractor
rather than \texttt{refex}, so constitutional and Bavarian \emph{Art.}\ citations
are included}), classified by jurisdiction and checked against the two knowledge bases. \emph{Federal} is the \texttt{gesetze-im-internet} dump the deployed pipeline searched; \emph{+ Bavarian} adds the 7{,}008 articles of Bavarian state law we collected from BAYERN.RECHT. A norm absent from the index cannot be retrieved however good the search is, so these figures bound recall from above.}
\label{tab:reference_coverage}
\small
\begin{tabular}{lrrrr}
\toprule
& \multicolumn{2}{c}{Citations} & \multicolumn{2}{c}{Present in corpus} \\
\cmidrule(lr){2-3}\cmidrule(lr){4-5}
Jurisdiction & $n$ & Share & Federal & + Bavarian \\
\midrule
Federal (Bundesrecht) & 2,067 & 49.2\% & 99.8\% & 99.8\% \\
State (Landesrecht) & 1,626 & 38.7\% & 0.0\% & 94.8\% \\
EU law & 69 & 1.6\% & 7.2\% & 7.2\% \\
Unattributed & 440 & 10.5\% & 48.0\% & 51.6\% \\
\midrule
\textbf{All} & \textbf{4,202} & \textbf{100.0\%} & \textbf{54.2\%} & \textbf{91.3\%} \\
\bottomrule
\end{tabular}
\end{table}

An ablation holding the pipeline fixed and swapping only the corpus (\Cref{tab:corpus}) recovers that ceiling without converting it: state-law recall@$10$  rises from $0$, which the federal dump forces by construction, to $4.31\%$, and gold-norm recall@$10$ from $2.28\%$ to $3.50\%$.
However, over the five generators that ran both corpora under the same RRF arm, the essay-score differences span $-0.80$ to $+1.79$ points and only DeepSeek-V4-Flash gains a point, while Gemma-4-31B loses ground. Therefore, adding Bavarian statutes helps retrieval coverage, but does not consistently improve downstream essay quality.

\begin{table}[ht]
  \centering
  \caption{What the Bavarian half of the corpus is worth. The retrieval rows hold the pipeline at RRF, so they read at the same setting as~\Cref{tab:ladder}; the essay rows are the round-robin arms, the only ones generated over the federal corpus. The federal corpus contains no state law, so its state-law recall is zero by construction while its overall recall is barely distinguishable -- overall recall is the wrong number to report here. Rows and shading follow~\Cref{tab:ladder}; the baseline column is the federal corpus.}
  \small
  \setlength{\tabcolsep}{3pt}
  \renewcommand{\arraystretch}{1.15}
  \begin{tabular}{lrr}
    \toprule
     & federal &  \\
     & only & combined \\
    \midrule
    \multicolumn{3}{l}{\emph{Retrieval quality (\%)}, against federal only} \\
    \quad gold-norm recall@10 & 2.28 & \cellcolor[HTML]{C5DFEC}3.50 \\
    \quad state-law recall@10 & 0.00 & \cellcolor[HTML]{C5DFEC}4.31 \\
    \midrule
    \multicolumn{3}{l}{\emph{Essay score (0--100)}, against federal only} \\
    \quad DeepSeek-V4-Flash & 43.33 & \cellcolor[HTML]{7EB8D7}45.12 \\
    \quad Gemma-4-31B-it & 34.57 & \cellcolor[HTML]{FDDBC7}33.77 \\
    \quad Qwen3-Next-80B-A3B & 29.63 & \cellcolor[HTML]{E6EFF4}30.00 \\
    \quad Qwen3.6-35B-A3B & 31.73 & \cellcolor[HTML]{E3EDF3}32.16 \\
    \quad gpt-oss-120B & 28.64 & \cellcolor[HTML]{DAE9F2}29.26 \\
    \bottomrule
  \end{tabular}
  \label{tab:corpus}
\end{table}

\Cref{tab:ladder} compares the arms on retrieval quality, essay score and citation coverage.
Gold-norm recall@$k$ rises nearly threefold across the eight search arms, from $3.50\%$ to $9.61\%$ of a ${\approx}19\%$ ceiling at $k{=}10$ (precision@$10$ of $50\%$) and $8.52\%$ to $23.92\%$ at $k=50$, while the largest gain any of them buys is $4.13$ points - DeepSeek-V4-Flash at \texttt{cite+LS+RRF}, $47.90$ (tying its oracle score) against $43.77$ - and no other generator gains more than $3.4$.
In short, gold-norm recall roughly triples across the retrieval methods, but the largest corresponding improvement in essay score is only $4.13$ points. Thus, substantial gains in retrieval quality do not translate into corresponding improvements in downstream legal-writing quality.

\begin{table}[ht]
  \centering
  \caption{RAG approaches for \gpbam{} Essay Writing:  \emph{RRF} for dense-hybrid rank-fusion; \emph{re-rank} reorders that pool with Qwen3-Reranker-4B; \emph{cite-only} resolves norms a rewriter suggests; \emph{cite+RRF} fills the remaining budget behind them; \emph{cite+LS+RRF} also resolves the Normenketten of the retrieved Leitsätze, \emph{LS text} provides their text.
  The \emph{oracle} supplies the provisions the reference solution cites, a coverage ceiling, not a RAG system.
  The last block notes the share of the solution's norms that the generated essay cites too.
  Shading is the difference from the block's own baseline, blue is better; bold is a paired difference beyond twice its standard error, uncorrected across cells. Essay scores are 3-judge medians.}
  \scriptsize
  \setlength{\tabcolsep}{3pt}
  \renewcommand{\arraystretch}{.95}
  \resizebox{\linewidth}{!}{%
  \begin{tabular}{lrrrrr|rrrr|r}
    \toprule
     & no &  &  & cite- & cite+ & RRF & cite+RRF & LS &cite+LS+ &  \\
     & RAG & RRF & rerank & only & RRF & $_{k=50}$ & $_{k{=}50}$ & text & RRF$_{k{=}50}$ & oracle \\
    \midrule
    \multicolumn{10}{l}{\quad  \emph{Retrieval quality (\%)}, against RRF} \\
    gold-norm recall@$k$ & -- & 3.50 & \cellcolor[HTML]{EFF3F5}5.33 & \cellcolor[HTML]{DEEBF2}9.04 & \cellcolor[HTML]{DBEAF2}9.61 & \cellcolor[HTML]{E0ECF3}8.52 & \cellcolor[HTML]{AED3E6}16.49 & \cellcolor[HTML]{EAF1F5}6.33 & \cellcolor[HTML]{71B0D3}23.92 & \cellcolor[HTML]{053061}\textcolor{white}{91.31} \\
     state-law recall@$k$ & -- & 4.31 & \cellcolor[HTML]{EDF2F5}6.15 & \cellcolor[HTML]{F8F3F0}3.26 & \cellcolor[HTML]{F6F7F7}4.31 & \cellcolor[HTML]{CFE4EF}11.62 & \cellcolor[HTML]{C2DDEC}12.98 & \cellcolor[HTML]{F2F5F6}5.17 & \cellcolor[HTML]{8AC0DB}19.07 & \cellcolor[HTML]{053061}\textcolor{white}{94.90} \\
     cases with a hit@$k$ & -- & 80.2 & \cellcolor[HTML]{A0CCE2}95.1 & \cellcolor[HTML]{96C7DF}96.3 & \cellcolor[HTML]{81BAD8}98.8 & \cellcolor[HTML]{8AC0DB}97.5 & \cellcolor[HTML]{75B2D4}100.0 & \cellcolor[HTML]{C5DFEC}90.1 & \cellcolor[HTML]{75B2D4}100.0 & \cellcolor[HTML]{75B2D4}100.0 \\
    \midrule
    \multicolumn{10}{l}{\quad  \emph{Essay score (0--100)}, against no RAG} \\
    DeepSeek-V4-Flash & 43.77 & \cellcolor[HTML]{DAE9F2}45.06 & \cellcolor[HTML]{F6F7F7}43.83 & \cellcolor[HTML]{E7F0F4}44.44 & \cellcolor[HTML]{FAE7DC}42.78 & \cellcolor[HTML]{E0ECF3}44.81 & \cellcolor[HTML]{F9EDE5}43.15 & \cellcolor[HTML]{FAE9DF}42.96 & \cellcolor[HTML]{6EAED2}\textbf{47.90} & \cellcolor[HTML]{6EAED2}\textbf{47.90} \\
     Gemma-4-31B & 32.90 & \cellcolor[HTML]{E0ECF3}33.95 & \cellcolor[HTML]{E0ECF3}33.95 & \cellcolor[HTML]{C5DFEC}\textbf{34.94} & \cellcolor[HTML]{93C6DE}\textbf{36.30} & \cellcolor[HTML]{E6EFF4}33.64 & \cellcolor[HTML]{BBDAEA}\textbf{35.19} & \cellcolor[HTML]{C5DFEC}34.94 & \cellcolor[HTML]{F9EBE3}32.22 & \cellcolor[HTML]{3E8CBF}\textbf{38.27} \\
     Qwen3-Next-80B-A3B & 30.62 & \cellcolor[HTML]{FBCEB7}28.52 & \cellcolor[HTML]{FAE8DE}29.69 & \cellcolor[HTML]{C5DFEC}32.65 & \cellcolor[HTML]{E1EDF3}31.60 & \cellcolor[HTML]{FBE5D8}29.51 & \cellcolor[HTML]{F6F7F7}30.62 & \cellcolor[HTML]{EFF3F5}30.99 & \cellcolor[HTML]{E9F0F4}31.23 & \cellcolor[HTML]{BDDBEA}32.84 \\
     Qwen3.6-35B-A3B & 30.62 & \cellcolor[HTML]{E4EEF4}31.48 & \cellcolor[HTML]{C2DDEC}32.72 & \cellcolor[HTML]{B8D8E9}32.96 & \cellcolor[HTML]{EDF2F5}31.05 & \cellcolor[HTML]{C7E0ED}32.59 & \cellcolor[HTML]{B6D7E8}33.09 & \cellcolor[HTML]{D8E9F1}31.98 & \cellcolor[HTML]{B6D7E8}\textbf{33.09} & \cellcolor[HTML]{569FC9}\textbf{35.31} \\
     gpt-oss-120B & 29.75 & \cellcolor[HTML]{F9EFE9}29.26 & \cellcolor[HTML]{FAE9DF}28.89 & \cellcolor[HTML]{B1D5E7}\textbf{32.35} & \cellcolor[HTML]{B1D5E7}32.35 & \cellcolor[HTML]{FCE2D2}28.46 & \cellcolor[HTML]{E6EFF4}30.49 & \cellcolor[HTML]{ECF2F5}30.25 & \cellcolor[HTML]{E4EEF4}30.62 & \cellcolor[HTML]{3480B9}\textbf{35.62} \\
     Magistral-Small & 23.33 & \cellcolor[HTML]{FCD7C2}\textbf{21.48} & \cellcolor[HTML]{FCE2D2}22.04 & \cellcolor[HTML]{FAE9DF}22.47 & \cellcolor[HTML]{FAE9DF}22.47 & \cellcolor[HTML]{FAEAE1}22.59 & \cellcolor[HTML]{F9EFE9}22.84 & \cellcolor[HTML]{FCE0D0}21.98 & \cellcolor[HTML]{F9EDE5}22.72 & \cellcolor[HTML]{B1D5E7}\textbf{25.93} \\
     Mistral-Small-3.2 & 21.98 & \cellcolor[HTML]{F9C4A9}\textbf{19.51} & \cellcolor[HTML]{FAEAE1}21.23 & \cellcolor[HTML]{F6F7F7}21.98 & \cellcolor[HTML]{F9EFE9}21.48 & \cellcolor[HTML]{E9F0F4}22.59 & \cellcolor[HTML]{FAE7DC}20.99 & \cellcolor[HTML]{E3EDF3}22.90 & \cellcolor[HTML]{E3EDF3}22.90 & \cellcolor[HTML]{9BC9E0}\textbf{25.19} \\
    \midrule
    \multicolumn{10}{l}{\quad  \emph{Gold norms cited by the essay (\%)}, against no RAG} \\
     DeepSeek-V4-Flash & 25.86 & \cellcolor[HTML]{EDF2F5}26.36 & \cellcolor[HTML]{E0ECF3}27.06 & \cellcolor[HTML]{F6F7F7}25.93 & \cellcolor[HTML]{F8F1ED}25.44 & \cellcolor[HTML]{D7E8F1}\textbf{27.48} & \cellcolor[HTML]{C2DDEC}\textbf{28.24} &  \cellcolor[HTML]{E4EEF4}26.82 & \cellcolor[HTML]{A5CEE3}\textbf{29.11} & \cellcolor[HTML]{053061}\textcolor{white}{\textbf{41.36}} \\
     Gemma-4-31B & 11.15 & \cellcolor[HTML]{E9F0F4}11.85 & \cellcolor[HTML]{E0ECF3}12.32 & \cellcolor[HTML]{E3EDF3}\textbf{12.16} & \cellcolor[HTML]{E0ECF3}\textbf{12.27} & \cellcolor[HTML]{CFE4EF}\textbf{13.16} & \cellcolor[HTML]{B3D6E8}\textbf{13.96} & \cellcolor[HTML]{D8E9F1}\textbf{12.65} & \cellcolor[HTML]{9DCBE1}\textbf{14.59} & \cellcolor[HTML]{053061}\textcolor{white}{\textbf{30.57}} \\
     Qwen3-Next-80B-A3B & 14.64 & \cellcolor[HTML]{D5E7F1}\textbf{16.33} & \cellcolor[HTML]{DAE9F2}\textbf{16.06} & \cellcolor[HTML]{D1E5F0}\textbf{16.57} & \cellcolor[HTML]{CCE2EF}\textbf{16.72} & \cellcolor[HTML]{B1D5E7}\textbf{17.50} & \cellcolor[HTML]{A9D1E5}\textbf{17.76} & \cellcolor[HTML]{D2E6F0}\textbf{16.44} & \cellcolor[HTML]{3A87BD}\textbf{20.86} & \cellcolor[HTML]{053061}\textcolor{white}{\textbf{40.59}} \\
     Qwen3.6-35B-A3B & 17.05 & \cellcolor[HTML]{DEEBF2}\textbf{18.28} & \cellcolor[HTML]{CCE2EF}\textbf{19.12} & \cellcolor[HTML]{E3EDF3}18.02 & \cellcolor[HTML]{DBEAF2}\textbf{18.40} & \cellcolor[HTML]{A2CDE3}\textbf{20.34} & \cellcolor[HTML]{BBDAEA}\textbf{19.65} & \cellcolor[HTML]{D4E6F1}\textbf{18.79} & \cellcolor[HTML]{8DC2DC}\textbf{20.97} & \cellcolor[HTML]{053061}\textcolor{white}{\textbf{36.57}} \\
     gpt-oss-120B & 15.04 & \cellcolor[HTML]{F7F5F4}14.90 & \cellcolor[HTML]{F2F5F6}15.27 & \cellcolor[HTML]{F3F5F6}15.25 & \cellcolor[HTML]{DAE9F2}\textbf{16.48} & \cellcolor[HTML]{E1EDF3}16.11 & \cellcolor[HTML]{DAE9F2}\textbf{16.53} & \cellcolor[HTML]{E7F0F4}15.85 & \cellcolor[HTML]{78B4D5}\textbf{19.49} & \cellcolor[HTML]{053061}\textcolor{white}{\textbf{35.61}} \\
     Magistral-Small & 11.43 & \cellcolor[HTML]{F9EDE5}10.70 & \cellcolor[HTML]{F9EDE5}10.74 & \cellcolor[HTML]{F9F0EB}10.91 & \cellcolor[HTML]{F7F6F6}11.37 & \cellcolor[HTML]{F9EBE3}10.67 & \cellcolor[HTML]{EFF3F5}11.83 & \cellcolor[HTML]{F3F5F6}11.59 & \cellcolor[HTML]{DEEBF2}12.64 & \cellcolor[HTML]{175290}\textcolor{white}{\textbf{19.78}} \\
     Mistral-Small-3.2 & 10.71 & \cellcolor[HTML]{FDDDCB}\textbf{8.96} & \cellcolor[HTML]{FCE2D2}\textbf{9.28} & \cellcolor[HTML]{F9EBE3}9.89 & \cellcolor[HTML]{FBE3D4}\textbf{9.32} & \cellcolor[HTML]{FAEAE1}9.83 & \cellcolor[HTML]{F8F1ED}10.31 & \cellcolor[HTML]{F8F4F2}10.56 & \cellcolor[HTML]{E3EDF3}11.70 & \cellcolor[HTML]{134C87}\textcolor{white}{\textbf{19.26}} \\
    \bottomrule
  \end{tabular}
  }

  \label{tab:ladder}
\end{table}

Overall, the norm oracle significantly improves six of seven generators, but retrieval cannot always close that gap (\Cref{tab:ladder}).
The oracle raises the point estimate for all seven generators and significantly
improves six. However, it does not narrow the differences between generators:
the two weakest generators gain $3.21$ and $2.60$ points, below the
seven-generator mean gain of $4.01$, and the across-generator standard deviation
increases from $7.15$ without retrieval to $7.73$ with the oracle
(\Cref{tab:oracle-summary}).
The gain is the statute \emph{wording}, not leaked structure or norm names: for all three, the oracle's norms without text do not help (\Cref{tab:oracle-ablation}). Specialized approaches, like norm lookup through headnotes, improve retrieval and generation, but not for all generators (only one reaches oracle level), while supplying headnote text directly does not significantly improve.

\begin{table}[t]
\centering
\caption{Summary of the oracle effect on essay scores, derived from \Cref{tab:ladder}.}
\footnotesize
\begin{tabular}{lr}
\toprule
Oracle comparison & Result \\
\midrule
Generators with a positive point estimate & $7/7$ \\
Generators with a significant improvement & $6/7$ \\
\midrule
Mistral-Small-3.2 gain (weakest) & $+3.21$ \\
Magistral-Small gain (second weakest) & $+2.60$ \\
Mean gain across seven generators & $+4.01$ \\
\midrule
Across-generator SD without retrieval & $7.15$ \\
Across-generator SD with oracle & $7.73$ \\
\bottomrule
\end{tabular}
\label{tab:oracle-summary}
\end{table}

\begin{table}[t]
  \centering
  \caption{Oracle advantage decomposition, relative to \norag{}.
  Bold exceeds twice the standard error.
  \emph{Wording withheld} supplies norms without their text; padding with junk norms and keeping 10 most-cited references test dilution; \emph{+ uncov. named} additionally names the cited norms not in the corpus.
    \emph{Order shuffled} drops the ranking by citation frequency, not an argument order.
  }
  \scriptsize
  \setlength{\tabcolsep}{3.5pt}
  \renewcommand{\arraystretch}{.95}
  \begin{tabular}{lrrrrrrr}
    \toprule
     & no &  & + uncov. & order & + junk & wording & top-10 \\
     & retrieval & oracle & named & shuffled & norms & withheld & only \\
    \midrule
    DeepSeek-V4-Flash & 43.77 & \cellcolor[HTML]{ACD2E5}$\mathbf{+4.13}$ & \cellcolor[HTML]{96C7DF}$\mathbf{+5.06}$ & \cellcolor[HTML]{D7E8F1}$+2.22$ & \cellcolor[HTML]{A9D1E5}$\mathbf{+4.20}$ & \cellcolor[HTML]{F7F5F4}$-0.19$ & \cellcolor[HTML]{CCE2EF}$+2.84$ \\

    Gemma-4-31B & 32.90 & \cellcolor[HTML]{8DC2DC}$\mathbf{+5.37}$ & \cellcolor[HTML]{71B0D3}$\mathbf{+6.23}$ & \cellcolor[HTML]{90C4DD}$\mathbf{+5.25}$ & \cellcolor[HTML]{90C4DD}$\mathbf{+5.25}$ & \cellcolor[HTML]{EFF3F5}$+0.56$ & \cellcolor[HTML]{8DC2DC}$\mathbf{+5.31}$ \\
    gpt-oss-120B & 29.75 & \cellcolor[HTML]{8AC0DB}$\mathbf{+5.87}$ & \cellcolor[HTML]{81BAD8}$\mathbf{+5.74}$ & \cellcolor[HTML]{BDDBEA}$\mathbf{+3.46}$ & \cellcolor[HTML]{D8E9F1}$+2.04$ & \cellcolor[HTML]{F6F7F7}$+0.00$ & \cellcolor[HTML]{8AC0DB}$\mathbf{+5.43}$ \\
    \bottomrule
  \end{tabular}

  \label{tab:oracle-ablation}
\end{table}

Retrieval is not uniformly beneficial either (\Cref{tab:ladder}): under hybrid search Mistral-Small-3.2 falls by $2.47$, and Magistral-Small by $1.85$, compared to no retrieval.
Neither of them, nor Qwen3-Next-80B, improves significantly with any RAG setup, though Mistral-Small-3.2 and Magistral-Small still gain significantly from the oracle ($+3.21$, $+2.60$), so the law does help them.

\section{Reliability of the Evaluation}

\subsection{Inter-Judge Agreement}
\label{sec:judge_analysis}

\begin{figure}
    \centering
    \includegraphics[width=\linewidth]{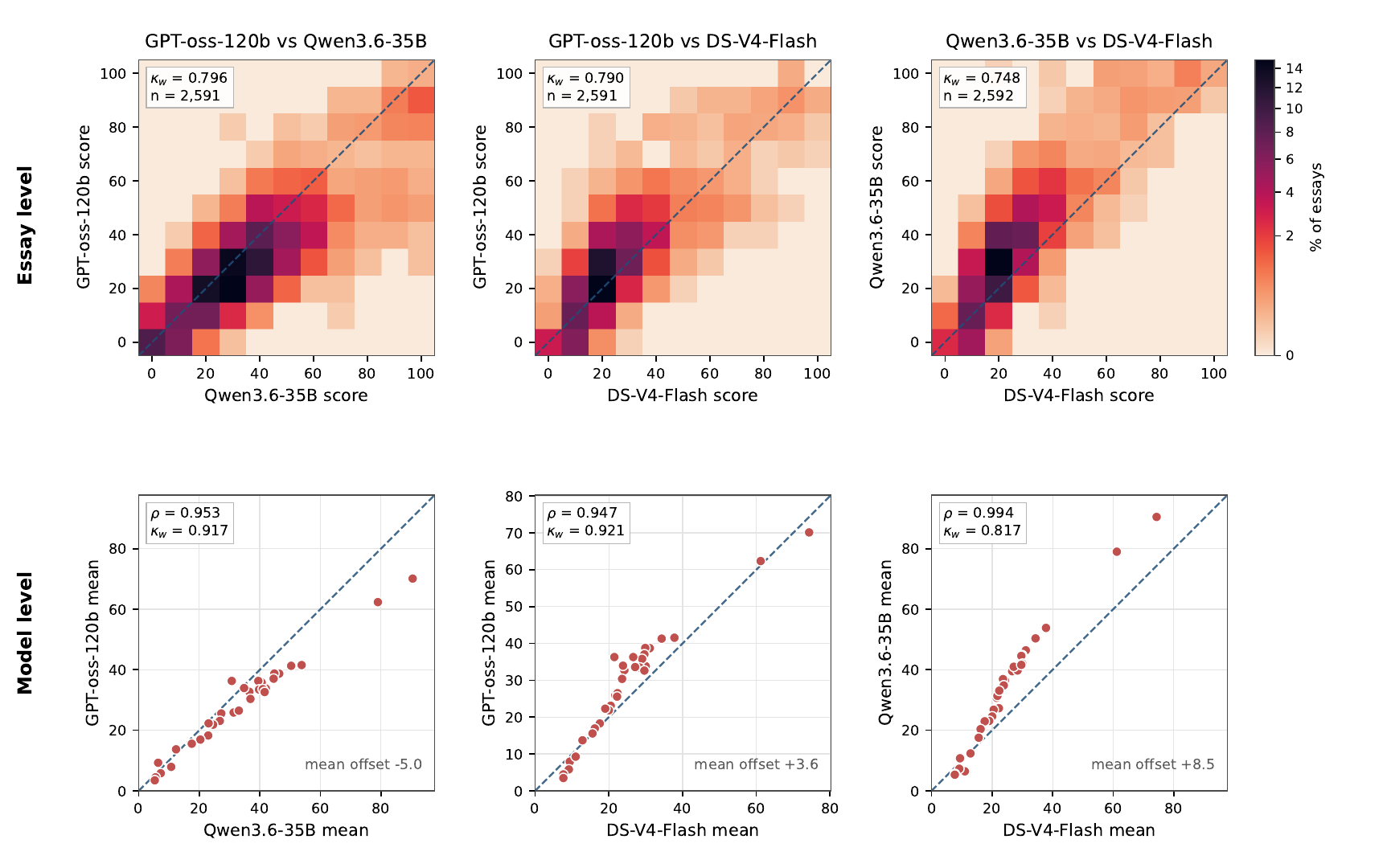}
    \caption{\textbf{Pairwise judge agreement without retrieval.}
Columns compare judge pairs. The top row shows joint essay-score distributions,
with colour indicating the percentage of essays; the bottom row compares each
judge's mean score per generation model. Dashed lines mark exact agreement.
Judges rank outputs similarly but differ systematically in score level.}
    \label{fig:inter-judge-aggrement-norag}
\end{figure}

\Cref{tab:judge_agreement_norag} reports pairwise agreement between the three
judge seats, measured both at model and at essay level, and
\Cref{fig:inter-judge-aggrement-norag} shows the score distributions those
coefficients summarise.

\paragraph{Rank agreement.} The three seats agree most strongly at the resolution the
benchmark reports. Over the 32 generation models every pair ranks alike at
$\rho{=}0.95$ or above; over individual essays the same pairs reach
$0.78$--$0.81$ ($n{=}2{,}591$--$2{,}592$). Averaging 81 essays per model cancels
much of the per-essay variation, so the model ordering in
\Cref{tab:main_results} is supported by stronger agreement than any individual
essay score is.

\paragraph{Level agreement.} Quadratic-weighted kappa separates the pairs
differently because it is sensitive to scoring severity whereas $\rho$ is not.
Qwen3.6-35B and DS-V4-Flash rank models most similarly ($\rho{=}0.99$), yet
show the lowest agreement in score level ($\kappa_w{=}0.82$ at model level and
$0.75$ at essay level). Qwen3.6-35B scores above DS-V4-Flash for 27 of the 32
models. Together with their high rank correlation, this indicates that the pair
differs primarily in severity rather than model ordering. The judges' mean
scores across the 32 models - $28.4$ for GPT-oss-120b, $33.4$ for Qwen3.6-35B,
and $24.8$ for DS-V4-Flash - show the same severity ordering visible in
\Cref{tab:main_results}. The three judges therefore largely agree on model
ranking and show substantial model-level agreement in absolute scores. Their
lower essay-level agreement is consistent with the greater variability of
individual essays compared with averages over 81 cases.

\begin{table}[htbp]
\centering
\caption{\textbf{Inter-judge agreement for no-retrieval essays.}
At model level, each observation is a generator's mean score; at essay level,
it is one essay. Spearman's $\rho$ measures rank agreement and
quadratic-weighted kappa ($\kappa_w$) agreement in score levels. Brackets show
95\% bootstrap CIs, clustered by generation model at essay level.}
\label{tab:judge_agreement_norag}
\setlength{\tabcolsep}{4pt}
\footnotesize
\resizebox{0.7\linewidth}{!}{%
\begin{tabular}{l l r r@{\hspace{2pt}}l r@{\hspace{2pt}}l}
\toprule
Level & Judge pair & $n$ & \multicolumn{2}{c}{Spearman $\rho$} & \multicolumn{2}{c}{QWK $\kappa_w$} \\
\midrule
Model & \shortstack[l]{GPT-oss-120b /\\ Qwen3.6-35B} & 32 & 0.95 & {\tiny$[0.84,\,0.99]$} & 0.92 & {\tiny$[0.87,\,0.94]$} \\
\addlinespace[5pt]
 & \shortstack[l]{GPT-oss-120b /\\ DS-V4-Flash} & 32 & 0.95 & {\tiny$[0.82,\,0.99]$} & 0.92 & {\tiny$[0.79,\,0.96]$} \\
\addlinespace[5pt]
 & \shortstack[l]{Qwen3.6-35B /\\ DS-V4-Flash} & 32 & 0.99 & {\tiny$[0.98,\,1.00]$} & 0.82 & {\tiny$[0.63,\,0.89]$} \\
\addlinespace[8pt]
Essay & \shortstack[l]{GPT-oss-120b /\\ Qwen3.6-35B} & 2591 & 0.81 & {\tiny$[0.71,\,0.86]$} & 0.80 & {\tiny$[0.71,\,0.84]$} \\
\addlinespace[5pt]
 & \shortstack[l]{GPT-oss-120b /\\ DS-V4-Flash} & 2591 & 0.78 & {\tiny$[0.68,\,0.84]$} & 0.79 & {\tiny$[0.64,\,0.86]$} \\
\addlinespace[5pt]
 & \shortstack[l]{Qwen3.6-35B /\\ DS-V4-Flash} & 2592 & 0.81 & {\tiny$[0.71,\,0.87]$} & 0.75 & {\tiny$[0.59,\,0.82]$} \\
\bottomrule
\end{tabular}
}
\end{table}

\subsection{A Strong External Judge}
\label{sec:opus_probe}

Inter-judge agreement measures consistency, not validity, and all three seats are mid-sized open-weight models.
As a partial check we had Claude-Opus-5 - well above every panel member on general capability - grade the 21 essays of the expert-annotation packet, seven cases crossed with three a priori capability tiers, using the byte-identical production request the panel seats received.
Two further items are blind duplicates judged by independent instances; both returned the same score, so the external judge's own  measurement noise is not what limits this comparison.

Opus-5 orders the essays much as the panel does.
Against the panel median it reaches $\rho{=}0.85$, and against the individual seats $0.65$--$0.92$; comparable to the $0.62$--$0.83$ agreement among the
panel judges on the same essays (\Cref{tab:opus_probe}).
Within a case, where the three essays differ only in the model that wrote them, mean Kendall $\tau_b$ is $+0.88$. The prespecified capability tiers have the same average ordering under both
Opus-5 and the panel median.

  \begin{table}[ht]
    \centering
    \small
\caption{\textbf{External-judge agreement on 21 essays.}
Opus-5--panel and within-panel agreement across seven cases and three
prespecified capability tiers. $\rho$ is Spearman correlation and $\kappa_w$
quadratic-weighted kappa.}
    \label{tab:opus_probe}
    \begin{tabular}{lrr}
      \toprule
      Pair & $\rho$ & $\kappa_w$ \\
      \midrule
      \multicolumn{3}{l}{\emph{Opus-5 against the panel}} \\
      \quad vs.\ panel median      & 0.85 & 0.65 \\
      \quad vs.\ gpt-oss-120b      & 0.92 & 0.64 \\
      \quad vs.\ Qwen3.6-35B-A3B   & 0.82 & 0.58 \\
      \quad vs.\ DeepSeek-V4-Flash & 0.65 & 0.67 \\
      \midrule
      \multicolumn{3}{l}{\emph{Panel internal, same 21 essays}} \\
      \quad gpt-oss-120b vs.\ Qwen3.6-35B-A3B      & 0.83 & 0.85 \\
      \quad gpt-oss-120b vs.\ DeepSeek-V4-Flash    & 0.62 & 0.55 \\
      \quad Qwen3.6-35B-A3B vs.\ DeepSeek-V4-Flash & 0.78 & 0.76 \\
      \bottomrule
    \end{tabular}
  \end{table}

Agreement on absolute level is weaker, and in one direction.
Regressing Opus-5 on the panel median gives a slope of $0.56$: the two agree almost exactly at the bottom
of the range and diverge as the panel's score rises.
The written verdicts locate the disagreement in fabricated norms and inverted results, which the panel appears to penalise less than a stronger judge does, rather than style - and that is the region the benchmark's spread lives in.
We report this as a bound on where the panel is most likely to be wrong, not as validation.
Opus-5 is one more judge with its own strictness prior, and no human grades exist for \gpbam{}.

\section{Discussion \& Conclusion}
\label{sec:discussion_conclusion}

We build \gpbam{} to answer what short-answer legal benchmarks cannot: ``\textit{When a model fails to write a \emph{Gutachten}, is it because it does not know the law, or because it cannot apply it?}'' 
The retrieval ladder  answers that by supplying the law and we find that the law helps, by a real but bounded amount.

\noindent\textit{Supplying the statute helps, but closes no gap.} 
An oracle supplying the provisions the solution cites is worth 
$2$--$6$ points across the seven generators (\Cref{tab:ladder})%
; handed that $91.3\%$, no generator cites even half of what it is given ($41.4\%$ at best).

\noindent\textit{Retrieval gains lag far behind recall.}
Gold-norm recall varies roughly threefold across our configurations while essay score moves by at most $4.1$ points, and 
the oracle's $91.31\%$ buys at most 3.3 more points over RAG.
How those norms are packaged matters far less consistently: padding, reordering and a ten-norm subset each cost some generators much of the gain and others almost none (\Cref{tab:oracle-ablation}).
That bears directly on work that evaluates the retrieval step in isolation~\cite{pipitone2024legalbenchragbenchmarkretrievalaugmentedgeneration}:
for these generators, a pipeline can be made much better at finding the right norms without becoming better at the task the norms are for.

\noindent\textit{The rewriter's memory outranks the search channel.} 
Naming the provisions and resolving them by exact key -- no search at all -- beats hybrid search on gold-norm recall, but not on state law (\Cref{tab:ladder}).
One reading is that statute text is simply the wrong register to search with a case: a \emph{Sachverhalt} raises doctrinal concepts, and a statute rarely names the doctrine it embodies. 
However, some models reproduce statutory text better than they write legal opinions
while the mid-tier proprietary systems do the reverse.
Perhaps they hold the material but cannot apply it in the \emph{Gutachtenstil} the examination rewards. 

\noindent\textit{The panel is internally consistent.} 
\looseness=-1
Three diverse judges  rank the 32 models near-identically ($\rho{=}0.95$--$0.99$)
so the ranking is not an artefact of any single seat. 
For our public-law cases, the difficulty of the
\textit{Gutachten} lies in applying law, not obtaining statutes -- which  a larger corpus cannot solve alone.

\section{Limitations \& Future Work}
\label{sec:limitations}
\textit{Limitation}. \gpbam{} contains 81 state-exam-style public-law cases from one Bavarian preparation course; results may not generalize to official examinations, other legal domains, institutions, or jurisdictions. 
All essay scores come from three LLM judges, without expert grades.
Their agreement measures consistency rather than validity and cannot detect shared bias.
Every RAG delta also carries a template change (\Cref{sec:essay_task}), so none isolates retrieval from framing.
The \emph{wording withheld} arm bounds it at $-0.19$ to $+0.56$ across all three (\Cref{tab:oracle-ablation}), including the generator with the largest retrieval gain.
Claude-opus-5 alone ran with a harness system prompt; transcripts show no access beyond the prompt file, but the prompt may account for part of its lead.
The scores should not be read as German examination grades or used to assess students.
At model level, score also correlates with answer length ($\rho{=}0.71$), and our design cannot distinguish greater completeness from a preference for longer answers~\cite{zheng2023judgingllmasajudgemtbenchchatbot,GU2026101253}.
The material is held out rather than released, so results cannot be independently reproduced from the sources, except for the released retrieval task.
Finally, the reported model-level correlations are descriptive and do not establish that statutory recall or citation overlap \textit{causes} better legal  reasoning.
The corpus is Bavarian and federal only: it excludes the 1{,}476 Verwaltungsvorschriften and holds just 7.2\% of the EU provisions the solutions cite.
It holds statutes in their current version%
, so a provision may have been amended since the reference solution was written.
All three judges are also generators, though the per-case median limits bias effects.

\label{sec:future_work}
\textit{Future work.} The judge panel should be validated against independent grades
from qualified examiners for \gpbam{} outputs  while accounting for variation in German exam grading~\cite{2024_hufeld_jede_korre}
and to establish whether the judge panel sits inside the range of human disagreement or outside it.
Our retrieval findings suggest that statutes are the wrong register to search with: in our corpus, doctrinal vocabulary of the kind a \emph{Sachverhalt} raises 
(e.g. \emph{Rücksichtnahmegebot}, \emph{Störerauswahl})
occurs only in 0.39\% of statute passages, while 38.5\% of court decisions use those terms.
\emph{Leitsätze} supply that missing register.
Our highest-recall arm already routes through them, resolving each retrieved headnote's \emph{Normenkette} rather than reading its text (\Cref{tab:ladder}), so searching the court decisions behind them
and integrating full commentaries are the next steps.

\section*{Acknowledgements}
\label{sec:acknowledgements}

The paper has been funded by (SCCH Comet) COMET K1-Competence Center for Integrated Software and AI Systems (INTEGRATE) within the Austrian COMET Program. 
This work was supported by the project FrAInderl, which is co-funded by the European Union.
We thank the Erlangen National High Performance Computing Center (NHR@FAU) for their LLM-as-a-Service offering.

\section*{GenAI Disclosure}
GenAI was used during the preparation of this manuscript to assist with code development and with drafting and revising text and tables to improve clarity and readability. All GenAI-assisted outputs were carefully reviewed, verified, edited, and incorporated by the authors as appropriate, who take full responsibility for the final manuscript.

\bibliographystyle{vancouver}
\bibliography{references} 

\appendix
\section{Additional Material}
\label{appendix:gpbam}

\subsection{Price and Performance}
\label{sec:price_performance}
\begin{figure}[htbp]
  \centering
  \includegraphics[width=.8\linewidth]{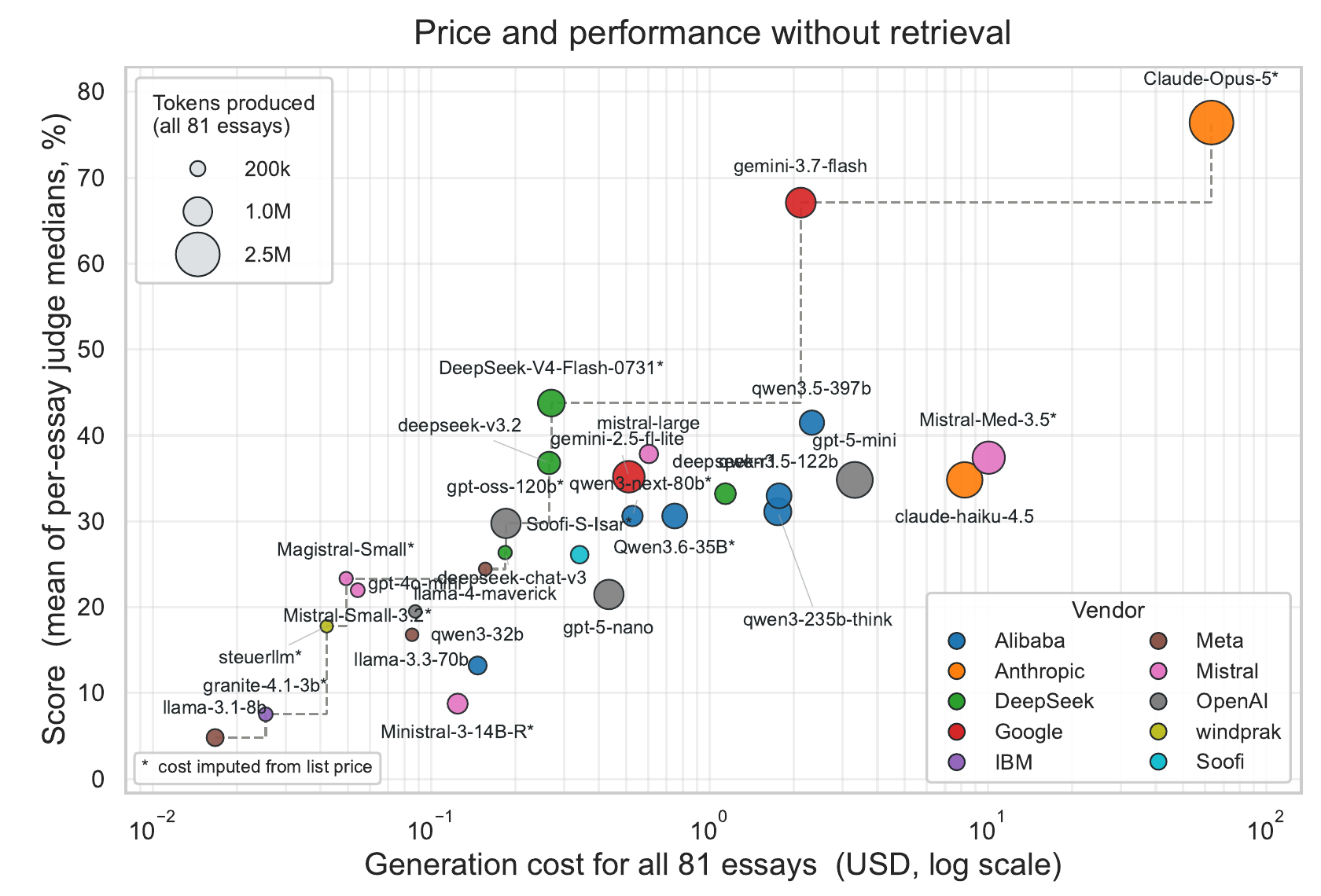}
  \caption{\textbf{GPBam Essay Writing price to performance.} One point per model,
  for the 29 paired models of Table~\ref{tab:main_results} with published prices.
  The horizontal axis is the full 81 generation cost in USD (log scale);
  the vertical axis is the essay writing \textbf{Score}.
  Bubble area is verbosity as the total number of  tokens the model produced over the 81 essays, colour is the vendor family.
  The dashed step line is the price/performance frontier.
  Models marked \texttt{*} were served self-hosted  and are instead priced at OpenRouter list rates
  (retrieved 2026-08-17).}
  \label{fig:cost_performance_norag}
  \end{figure}

\Cref{tab:main_results} reports what each model achieves on the essay task while \Cref{fig:cost_performance_norag} reports on the cost dimension for the \norag{} condition.

\Cref{fig:cost_performance_norag} plots each model's essay score against the
dollar cost of generating all 81 opinions. The two are far from proportional:
cost ranges over four orders of magnitude while scores vary by a factor of
sixteen, and the price--performance frontier advances in a small number of
discrete steps rather than smoothly. Absolute expenditure is nonetheless
modest. At the list prices used here, the complete no-retrieval study projects
to \$99 for generation, \$32 for judging and \$25 for article recitation
(\Cref{tab:main_results}).

\subsection{Prompts}
\label{app:all_prompt}

All prompts are reproduced verbatim from the code that sent them.
\emph{The German wording is the only one that was ever sent to a model.} Each
German block is followed by an English translation for the reader, which was
used in no run. The translation keeps the layout of the German line for line --
same line numbers, blank lines, indentation, list markers and slots -- so that
line~$n$ of the English translates line~$n$ of the German. Strings that a model
received literally or that the score parser matches (XML tags, attribute names,
the score format) are left untranslated. \texttt{\{\}} marks a slot filled per
call. Line numbers count lines of the prompt; a $\hookrightarrow$ marks where a
long line is wrapped for print and is not part of the prompt.

\subsubsection{\norag{}: Question Answer Prompt}
\label{appendix:qa_prompt}

Every legal opinion scored in this paper is produced by this prompt or by its
\lawrag{} variant (\Cref{appendix:qa_rag_prompt}). The request consists of
the user message below and an empty system message; the one exception is
Claude-opus-5, which ran through an agent harness that supplies its own system
prompt. The user message has two slots: the context slot (line~17), which is
empty under \norag{}, and the \emph{Sachverhalt} of one case
(line~21). No reference solution enters the prompt. The shaded lines 11, 13
and 14 occur only under \norag{}; the \lawrag{} prompt drops them
(\Cref{appendix:qa_rag_prompt}).

\smallskip
\needspace{5\baselineskip}
\noindent\textbf{User message (German, as sent)}\par\nopagebreak
\begin{prompt}[highlightlines={11,13-14}]
Du bist ein Experte für deutsches öffentliches Recht und bearbeitest Rechtsfragen in strukturierter Weise im Stil eines juristischen Gutachtens.
Du sollst ein Rechtsgutachten zu den dargestellten Rechtsfällen erstellen, wie in der Frage angewiesen.
Es handelt sich um eine juristische Examensklausur auf Staatsexamensniveau, für deren Beantwortung den Studierenden 4 Stunden zur Verfügung stehen, mit einem ausführlichen Inhaltsverzeichnis, gefolgt von einer vollständigen juristischen Würdigung.
Gehe davon aus, dass ausschließlich deutsches Recht anwendbar ist, sofern nicht ausdrücklich etwas anderes erwähnt wird.
Verwende präzise juristische Sprache und antworte so gründlich wie möglich.
Gib KEINEN Haftungsausschluss ab und verweise nicht auf die Notwendigkeit externer Rechtsberatung.
Fordere den Nutzer NICHT auf, Gesetze zu konsultieren oder eigene Recherchen anzustellen.
Sprich direkt und autoritativ, ohne zu erwähnen, dass deine Antwort lediglich allgemeinen Informationszwecken dient.
Integriere deutschlandspezifische juristische Terminologie.
Wenn du relevante rechtliche Erwägungen entdeckt hast, antworte mit einer knappen, klaren juristischen Analyse.
Zitiere ausschließlich aus den von dir identifizierten Erwägungen.
Zitiere stets die konkrete Rechtsnorm und gib ausdrücklich Absätze (Abs.), Sätze (S.), Nummern (Nr.) oder Buchstaben (Buchst.) an, soweit vorhanden, z. B. „Art. 12 Abs. 1 GG“, „§ 35 Satz 1 VwVfG“, „§ 42 Abs. 2 VwGO“, „§ 823 Abs. 1 BGB“ oder „§ 263 Abs. 1 StGB“. Vermeide allgemeine Verweise wie „Art. 3 GG“ oder „§ 242 BGB“, ohne den konkreten Absatz, Satz, die Nummer oder den Buchstaben zu nennen, sofern einschlägig.
Wenn keine relevanten Erwägungen gefunden werden, stelle ausdrücklich fest, dass keine einschlägigen Informationen verfügbar sind.
Wenn dir verlässliche Quellen vorliegen, teile daraus praktische Hinweise oder Erkenntnisse mit.
Wenn die Eingabe verlangt, einen konkreten in der Klausur angegebenen Fall zu analysieren, der Text oder die Einzelheiten dieses Falls jedoch in der Eingabe nicht bereitgestellt wurden, weise ausdrücklich darauf hin, dass das erforderliche Fallmaterial fehlt.

{}

Frage:
```
{}
```

Antwort:
\end{prompt}

\needspace{5\baselineskip}
\noindent\textbf{User message (English translation, not used in any run)}\par\nopagebreak
\begin{prompt}[highlightlines={11,13-14}]
You are an expert in German public law and treat legal questions in a structured way in the style of a legal opinion.
You are to produce a legal opinion on the cases presented, as directed in the question.
This is a law examination at state-examination level, for which students have 4 hours, with a detailed table of contents followed by a complete legal assessment.
Assume that only German law applies unless something else is expressly mentioned.
Use precise legal language and answer as thoroughly as possible.
Do NOT give a disclaimer and do not refer to the need for external legal advice.
Do NOT ask the user to consult statutes or to carry out their own research.
Speak directly and authoritatively, without mentioning that your answer serves merely general informational purposes.
Incorporate Germany-specific legal terminology.
If you have identified relevant legal considerations, answer with a concise, clear legal analysis.
Cite exclusively from the considerations you have identified.
Always cite the specific legal provision and state expressly the paragraphs (Abs.), sentences (S.), numbers (Nr.) or letters (Buchst.) where these exist, e.g. "Art. 12 Abs. 1 GG", "§ 35 Satz 1 VwVfG", "§ 42 Abs. 2 VwGO", "§ 823 Abs. 1 BGB" or "§ 263 Abs. 1 StGB". Avoid general references such as "Art. 3 GG" or "§ 242 BGB" without naming the specific paragraph, sentence, number or letter, where applicable.
If no relevant considerations are found, state expressly that no pertinent information is available.
If reliable sources are available to you, share practical guidance or findings from them.
If the input requires analysis of a specific case stated in the examination but the text or details of that case were not provided in the input, point out expressly that the required case material is missing.

{}

Question:
```
{}
```

Answer:
\end{prompt}

\subsubsection{\lawrag{}: Question Answer Prompt}
\label{appendix:qa_rag_prompt}

The \lawrag{} user message differs from the \norag{} one in two places.
First, three lines of the \norag{} prompt are dropped (shaded lines 11, 13 and
14 in \Cref{appendix:qa_prompt}): \emph{``Zitiere ausschließlich aus den von
dir identifizierten Erwägungen.''}, \emph{``Wenn keine relevanten Erwägungen
gefunden werden, \ldots''} and \emph{``Wenn dir verlässliche Quellen vorliegen,
\ldots''}. Second, a block on how to read the retrieved statutes, shaded as
lines 14--19 below, is inserted before the context slot. The system message is again empty. The three dropped lines were written
for a generic retrieval assistant: instructing a model to cite exclusively from
the considerations it identified, or to state that no pertinent information is
available, does not fit an answer written over a \texttt{<context>} block --
and the \norag{} condition inherits them because it is the same template with
an empty slot. The two conditions therefore differ in their instructions as
well as in their context. The context slot is line~21, the
\emph{Sachverhalt} line~25.

\smallskip
\needspace{5\baselineskip}
\noindent\textbf{User message (German, as sent)}\par\nopagebreak
\begin{prompt}[highlightlines={14-19}]
Du bist ein Experte für deutsches öffentliches Recht und bearbeitest Rechtsfragen in strukturierter Weise im Stil eines juristischen Gutachtens.
Du sollst ein Rechtsgutachten zu den dargestellten Rechtsfällen erstellen, wie in der Frage angewiesen.
Es handelt sich um eine juristische Examensklausur auf Staatsexamensniveau, für deren Beantwortung den Studierenden 4 Stunden zur Verfügung stehen, mit einem ausführlichen Inhaltsverzeichnis, gefolgt von einer vollständigen juristischen Würdigung.
Gehe davon aus, dass ausschließlich deutsches Recht anwendbar ist, sofern nicht ausdrücklich etwas anderes erwähnt wird.
Verwende präzise juristische Sprache und antworte so gründlich wie möglich.
Gib KEINEN Haftungsausschluss ab und verweise nicht auf die Notwendigkeit externer Rechtsberatung.
Fordere den Nutzer NICHT auf, Gesetze zu konsultieren oder eigene Recherchen anzustellen.
Sprich direkt und autoritativ, ohne zu erwähnen, dass deine Antwort lediglich allgemeinen Informationszwecken dient.
Integriere deutschlandspezifische juristische Terminologie.
Wenn du relevante rechtliche Erwägungen entdeckt hast, antworte mit einer knappen, klaren juristischen Analyse.
Zitiere stets die konkrete Rechtsnorm und gib ausdrücklich Absätze (Abs.), Sätze (S.), Nummern (Nr.) oder Buchstaben (Buchst.) an, soweit vorhanden, z. B. „Art. 12 Abs. 1 GG“, „§ 35 Satz 1 VwVfG“, „§ 42 Abs. 2 VwGO“, „§ 823 Abs. 1 BGB“ oder „§ 263 Abs. 1 StGB“. Vermeide allgemeine Verweise wie „Art. 3 GG“ oder „§ 242 BGB“, ohne den konkreten Absatz, Satz, die Nummer oder den Buchstaben zu nennen, sofern einschlägig.
Wenn die Eingabe verlangt, einen konkreten in der Klausur angegebenen Fall zu analysieren, der Text oder die Einzelheiten dieses Falls jedoch in der Eingabe nicht bereitgestellt wurden, weise ausdrücklich darauf hin, dass das erforderliche Fallmaterial fehlt.

Im Abschnitt <context> findest du Normtexte, die durch eine automatische Suche zu diesem Sachverhalt gefunden wurden. Beachte dabei:
- Die Auswahl ist unvollständig und teilweise unpassend. Sie ersetzt nicht deine eigene Kenntnis des deutschen Rechts.
- Insbesondere kann Landesrecht (z. B. PAG, BayBO, BayVwVfG, BV, GO, LStVG) vollständig fehlen, obwohl es einschlägig ist.
- Wenn eine Vorschrift dort nicht auftaucht, heißt das nicht, dass sie nicht existiert oder nicht einschlägig ist. Zitiere die einschlägigen Normen unabhängig davon, ob ihr Wortlaut unten abgedruckt ist.
- Verwende die abgedruckten Normtexte, soweit sie einschlägig sind, und ignoriere die übrigen. Übernimm keine Vorschrift nur deshalb, weil sie unten steht.
- Die Normtexte stehen in <norm>-Elementen mit den Attributen zitat, gebiet und ueberschrift. Zitiere im Gutachten die Norm selbst (etwa „§ 34 BauGB“), nicht das Element oder seine Position.

{}

Frage:
```
{}
```

Antwort:
\end{prompt}

\needspace{5\baselineskip}
\noindent\textbf{User message (English translation, not used in any run)}\par\nopagebreak
\begin{prompt}[highlightlines={14-19}]
You are an expert in German public law and treat legal questions in a structured way in the style of a legal opinion.
You are to produce a legal opinion on the cases presented, as directed in the question.
This is a law examination at state-examination level, for which students have 4 hours, with a detailed table of contents followed by a complete legal assessment.
Assume that only German law applies unless something else is expressly mentioned.
Use precise legal language and answer as thoroughly as possible.
Do NOT give a disclaimer and do not refer to the need for external legal advice.
Do NOT ask the user to consult statutes or to carry out their own research.
Speak directly and authoritatively, without mentioning that your answer serves merely general informational purposes.
Incorporate Germany-specific legal terminology.
If you have identified relevant legal considerations, answer with a concise, clear legal analysis.
Always cite the specific legal provision and state expressly the paragraphs (Abs.), sentences (S.), numbers (Nr.) or letters (Buchst.) where these exist, e.g. "Art. 12 Abs. 1 GG", "§ 35 Satz 1 VwVfG", "§ 42 Abs. 2 VwGO", "§ 823 Abs. 1 BGB" or "§ 263 Abs. 1 StGB". Avoid general references such as "Art. 3 GG" or "§ 242 BGB" without naming the specific paragraph, sentence, number or letter, where applicable.
If the input requires analysis of a specific case stated in the examination but the text or details of that case were not provided in the input, point out expressly that the required case material is missing.

In the <context> section you will find statutory texts found by an automatic search for this case. Note the following:
- The selection is incomplete and in part unsuitable. It does not replace your own knowledge of German law.
- In particular, state law (e.g. PAG, BayBO, BayVwVfG, BV, GO, LStVG) may be missing entirely even though it is applicable.
- If a provision does not appear there, that does not mean it does not exist or is not applicable. Cite the applicable provisions regardless of whether their wording is printed below.
- Use the printed statutory texts insofar as they are applicable, and ignore the rest. Do not adopt a provision merely because it appears below.
- The statutory texts appear in <norm> elements with the attributes zitat, gebiet and ueberschrift. In the opinion, cite the provision itself (such as "§ 34 BauGB"), not the element or its position.

{}

Question:
```
{}
```

Answer:
\end{prompt}

The context slot is filled with one \texttt{<norm>} element per retrieved
provision, carrying its conventional citation, its jurisdiction
(\texttt{Bundesrecht} or \texttt{Landesrecht Bayern}) and its official heading
as attributes and its statutory text as the element body. Attributes rather
than a prose header, so that the metadata cannot be mistaken for norm text, and
no running index, so that the model has no position number to echo instead of
a citation. Bracketed italics below are placeholders, not part of the prompt:

\begin{prompt}[numbers=none, commandchars=\\\{\}]
<context>
<norm zitat="§ 34 BauGB" gebiet="Bundesrecht" ueberschrift="Zulässigkeit von Vorhaben innerhalb der im Zusammenhang bebauten Ortsteile">
\textrm{\emph{[statutory text]}}
</norm>
\textrm{\emph{[further <norm> elements]}}
</context>
\end{prompt}

\subsubsection{Article Recitation Prompt}
\label{appendix:article_recitation_prompt}

Every Article Recitation answer scored in this paper is produced by the two
messages below; no case facts, retrieved passages or reference solution enter
the prompt. The prompt is written in German, matching the language of the
statutes it asks for. The single slot carries the provision identifier, filled from
whichever set the prompt is drawn from -- \texttt{vwgo 80} in \emph{GPBam laws},
\texttt{VwGO \S~154} in \emph{Most-cited laws}
(\Cref{sec:article_recitation_task}).

\smallskip
\needspace{5\baselineskip}
\noindent\textbf{System message (German, as sent)}\par\nopagebreak
\begin{prompt}
Du bist ein Rechercheur für deutsches Recht. Beantworte die folgenden Fragen und gib direkt den Inhalt der neuesten Fassung des Gesetzes wieder, ohne Warnungen, Rückfragen oder Einleitung. Falls du dazu nicht in der Lage bist, liefere die ausführlichste Zusammenfassung, die du geben kannst.
\end{prompt}

\needspace{5\baselineskip}
\noindent\textbf{User message (German, as sent)}\par\nopagebreak
\begin{prompt}
Wie lautet der genaue Wortlaut von {}?
\end{prompt}

\needspace{5\baselineskip}
\noindent\textbf{System message (English translation, not used in any run)}\par\nopagebreak
\begin{prompt}
You are a researcher of German law. Answer the following questions and reproduce directly the content of the most recent version of the statute, without warnings, follow-up questions or introduction. If you are unable to do so, supply the most detailed summary you can give.
\end{prompt}

\needspace{5\baselineskip}
\noindent\textbf{User message (English translation, not used in any run)}\par\nopagebreak
\begin{prompt}
What is the exact wording of {}?
\end{prompt}

\subsubsection{Judge Prompt}
\label{appendix:judge_prompt}

Every judge score reported in this paper is produced by the two messages below.
The prompt is written in German, matching the language of the cases, the
reference solutions and the generated essays. The three slots carry the case (line~34), its reference solution
(line~41) and the generated essay (line~47). The score is
read back from the mandatory \texttt{[[\,$\cdot$\,]]} marker; a response without
a parsable marker counts as missing and is skipped rather than imputed
(\Cref{sec:metrics}).

\smallskip
\needspace{5\baselineskip}
\noindent\textbf{System message (German, as sent)}\par\nopagebreak
\begin{prompt}
Handle als Prüfer, der auf die Bewertung universitärer Klausuren an deutschen rechtswissenschaftlichen Fakultäten spezialisiert ist.
Deine Aufgabe ist es, zu beurteilen, wie gut die Antwort mit der Musterlösung übereinstimmt, mit Schwerpunkt auf Richtigkeit, Vollständigkeit und juristischer Argumentation.
\end{prompt}

\needspace{5\baselineskip}
\noindent\textbf{User message (German, as sent)}\par\nopagebreak
\begin{prompt}
Bewerte den nachstehenden Fall, gib eine kurze Begründung und die endgültige Richtigkeitsbewertung an.

Ziel:
Deine Aufgabe ist es zu beurteilen, wie gut die Antwort mit der Referenzantwort übereinstimmt, wobei der Schwerpunkt auf Genauigkeit, Vollständigkeit und juristischer Argumentation liegt.
Die generierte Antwort muss alle Aspekte der Referenzantwort in ähnlicher Detailtiefe und mit demselben Ergebnis abdecken, um die volle Punktzahl zu erhalten. 
Wenn Punkte, die für die Referenzantwort wesentlich sind, nicht erörtert werden, sollte dies zu einem erheblichen Punktabzug führen, da dadurch die juristische Argumentation unterbrochen wird.
Die generierte Antwort sollte daher einer ähnlichen Struktur wie die Referenzantwort folgen. Jede Abweichung von der Referenzantwort muss im Hinblick auf die in der Referenzantwort gesetzten Maßstäbe präzise begründet werden.
Eine Antwort, die kurz ist, aber die Tiefe der Subsumtion aus der Referenzantwort auslässt, muss deutlich niedriger bewertet werden als eine lange Antwort, die eine Subsumtion versucht, selbst wenn die lange Antwort geringfügige stilistische Abweichungen enthält.
Daher ist in einem ersten Schritt die Referenzantwort in logisch zusammenhängende Inhaltsabschnitte aufzuteilen, und diese sind den entsprechenden Absätzen in der Antwort des Modells zuzuordnen. Gib für jeden dieser Absätze Feedback gemäß dem unten im Rückgabeformat beschriebenen Schema. Stelle sicher, dass alle Normzitate im selben Inhaltsbereich ebenfalls korrekt zitiert sind.

Kontext:
Dir wird eine vollständige juristische Antwort bereitgestellt (gekennzeichnet als: Antwort des Modells im <antwort_des_modells>-Tag), die auf einer juristischen Examensfrage (gekennzeichnet als: Frage im <frage>-Tag) und einer Musterlösung (gekennzeichnet als: Referenzantwort im <referenzantwort>-Tag) basiert.

Rückgabeformat:
    Nach der Überprüfung der Antwort:
    1. Erklärung: Erkläre deine Begründung dazu, inwiefern der generierte Aufsatz mit dem Referenzaufsatz übereinstimmt oder von ihm abweicht. 
        Identifiziere alle zentralen juristischen Schritte in der Referenzantwort wie oben beschrieben. Enthält die Antwort des Modells Schritt 1? Schritt 2?
    2. Konstruktives Feedback: Gib zusätzlich neutrales, konstruktives Feedback und Korrekturen im Stil eines Universitätsprofessors. Das Feedback sollte präzise und direkt aufzeigen, was falsch ist, und Hinweise zur Verbesserung geben.
    3. Korrektheitsbewertung: Vergib eine abschließende Korrektheitsbewertung auf einer Skala von 0,0 bis 1,0 (in Schritten von 0,1). Diese Bewertung sollte widerspiegeln, in welchem Umfang die Antwort die Referenzantwort erfüllt, wobei 
        - 1,0 = vollständige Erfüllung (100 %) 
        - niedrigere Werte verhältnismäßige Defizite widerspiegeln (z. B. 0,5 = 50 % Erfüllung). 
        - Halte dich strikt an das Format: "[[Punktzahl]]", z. B. "Die Richtigkeitsbewertung: [[0,5]]".

Warnhinweise:
    - Abweichungen oder zusätzliche Elemente, die in der Referenzantwort nicht enthalten sind, sollten unberücksichtigt bleiben, sofern du nicht sicher bist, dass sie juristisch korrekt und relevant sind. 
        Gehe davon aus, dass die Referenzantwort alle Informationen enthält, die für eine perfekte Antwort und Bewertung erforderlich sind.
    - Die Referenzantwort kann Zitate enthalten (z. B. aus Büchern oder juristischen Fachaufsätzen), die die Antwort nicht wiedergeben muss. 
        Gesetzesbestimmungen sollten jedoch präzise zitiert werden, wobei Abs., S., Nr. oder Buchst. anzugeben sind, sofern dies anwendbar ist.
    - Wenn die Referenzantwort separate Unterpunkte enthält, nutze diese als Orientierung für eine proportionale Bewertung (z. B. entspricht die korrekte Behandlung von 2 von 4 Unterpunkten ungefähr einer Bewertung von 0,5).

Frage:

<frage>
{}
</frage>


Referenzantwort:

<referenzantwort>
{}
</referenzantwort>

Antwort des Modells:

<antwort_des_modells>
{}
</antwort_des_modells>

Dein Urteil:
\end{prompt}

\needspace{5\baselineskip}
\noindent\textbf{System message (English translation, not used in any run)}\par\nopagebreak
\begin{prompt}
Act as an examiner specialised in grading university examinations at German law faculties.
Your task is to assess how well the answer agrees with the model solution, with emphasis on correctness, completeness and legal reasoning.
\end{prompt}

\needspace{5\baselineskip}
\noindent\textbf{User message (English translation, not used in any run)}\par\nopagebreak
\begin{prompt}
Grade the case below, give a brief justification and state the final correctness score.

Goal:
Your task is to assess how well the answer agrees with the reference answer, with the emphasis on accuracy, completeness and legal reasoning.
To receive full marks, the generated answer must cover every aspect of the reference answer at a similar level of detail and with the same outcome.
If points that are essential to the reference answer are not discussed, this should lead to a substantial deduction, because it breaks the chain of legal reasoning.
The generated answer should therefore follow a structure similar to the reference answer. Any deviation from the reference answer must be justified precisely against the standards the reference answer sets.
An answer that is short but omits the depth of subsumption found in the reference answer must be scored markedly lower than a long answer that attempts a subsumption, even if the long answer contains minor stylistic deviations.
As a first step, therefore, divide the reference answer into logically coherent content sections and map these onto the corresponding paragraphs of the model's answer. Give feedback on each of these paragraphs according to the scheme described under Return format below. Make sure that all statutory citations within the same content section are also cited correctly.

Context:
You are given a complete legal answer (marked as: the model's answer in the <antwort_des_modells> tag), which is based on a legal examination question (marked as: the question in the <frage> tag) and a model solution (marked as: the reference answer in the <referenzantwort> tag).

Return format:
    After reviewing the answer:
    1. Explanation: Explain your reasoning as to how far the generated essay agrees with or departs from the reference essay.
        Identify all the central legal steps in the reference answer as described above. Does the model's answer contain step 1? Step 2?
    2. Constructive feedback: In addition, give neutral, constructive feedback and corrections in the style of a university professor. The feedback should point out precisely and directly what is wrong and give guidance for improvement.
    3. Correctness score: Award a final correctness score on a scale from 0.0 to 1.0 (in steps of 0.1). This score should reflect the extent to which the answer fulfils the reference answer, where
        - 1.0 = complete fulfilment (100 %)
        - lower values reflect proportionate deficits (e.g. 0.5 = 50 % fulfilment).
        - Adhere strictly to the format: "[[Punktzahl]]", e.g. "Die Richtigkeitsbewertung: [[0,5]]".

Cautions:
    - Deviations or additional elements not contained in the reference answer should be disregarded unless you are certain that they are legally correct and relevant.
        Assume that the reference answer contains all the information required for a perfect answer and a perfect assessment.
    - The reference answer may contain citations (e.g. to books or legal journal articles) that the answer need not reproduce.
        Statutory provisions, however, should be cited precisely, giving Abs., S., Nr. or Buchst. where applicable.
    - If the reference answer contains separate sub-points, use these as a guide for proportional scoring (e.g. correctly handling 2 of 4 sub-points corresponds roughly to a score of 0.5).

Question:

<frage>
{}
</frage>


Reference answer:

<referenzantwort>
{}
</referenzantwort>

Model's answer:

<antwort_des_modells>
{}
</antwort_des_modells>

Your judgement:
\end{prompt}

\end{document}